%% file: submission.tex
\documentclass[10pt,twocolumn,letterpaper]{article}

\usepackage[pagenumbers]{cvpr} %

\usepackage{graphicx}
\usepackage{amsmath}
\usepackage{amssymb}
\usepackage{booktabs}

\usepackage[pagebackref,breaklinks,colorlinks]{hyperref}
\usepackage[hang,flushmargin]{footmisc}

\input{util/packages}
\input{util/commands}

\def\cvprPaperID{5990} %
\def\confName{CVPR}
\def\confYear{2022}

\begin{document}

\title{Task2Sim: Towards Effective Pre-training and Transfer from Synthetic Data}

\author{Samarth Mishra$^{\dagger1}$ \quad Rameswar Panda$^{2}$ \quad Cheng Perng Phoo$^{\dagger3}$ \quad Chun-Fu (Richard) Chen$^{*2}$ \\
\quad Leonid Karlinsky$^2$ \quad
Kate Saenko$^{1,2}$ \qquad Venkatesh Saligrama$^1$ \qquad Rogerio S. Feris$^{2}$ \\
$^1$Boston University \quad $^2$MIT-IBM Watson AI Lab \quad $^3$Cornell University 
}

\maketitle

\input{latex/abstract}

\blfootnote{$^\dagger$Work done as interns at MIT-IBM Watson AI Lab.}
\blfootnote{$^*$Now affiliated with JPMorgan Chase, FLARE. Work done when Chun-Fu was at MIT-IBM Watson AI Lab.}
\blfootnote{Project page : \href{https://samarth4149.github.io/projects/task2sim.html}{https://samarth4149.github.io/projects/task2sim.html}}

\vspace{-4mm}
\input{latex/introduction}

\input{latex/related_work}

\input{latex/method}

\input{latex/experiments}

\input{latex/conclusion}

{\small
\bibliographystyle{ieee_fullname}
\bibliography{submission}
}

\input{latex/appendix}

\end{document}

%% file: util/packages.tex
\usepackage{booktabs}       %
\usepackage{multirow}
\usepackage{amsmath,amssymb}
\usepackage{color}
\usepackage{caption}
\usepackage{subcaption}
\usepackage{graphicx}
\usepackage{xspace}
\usepackage{bbm}
\usepackage{makecell}
\usepackage{nccmath}
\usepackage{xcolor}
\usepackage{csquotes}

\usepackage[ruled, lined, longend, linesnumbered]{algorithm2e}
\usepackage[capitalize]{cleveref}

%% file: util/commands.tex
\newcommand{\mc}{\mathcal}

\newcommand{\bs}{\boldsymbol}

\newcommand{\name}{Task2Sim}
\def\ours{Task2Sim\xspace}

\DeclareMathOperator*{\argmax}{arg\,max}

\DeclareMathOperator*{\E}{\mathbb{E}}

\makeatletter
\DeclareRobustCommand\onedot{\futurelet\@let@token\@onedot}

\def\@onedot{\ifx\@let@token.\else.\null\fi\xspace}

\def\eg{\emph{e.g}\onedot} 
\def\ie{\emph{i.e}\onedot} 
 
\def\etc{\emph{etc}\onedot}

\newcommand\blfootnote[1]{%
  \begingroup
  \renewcommand\thefootnote{}\footnote{#1}%
  \addtocounter{footnote}{-1}%
  \endgroup
}

\Crefname{figure}{Figure}{Figures}
\Crefname{section}{Section}{Sections}
\Crefname{table}{Table}{Tables}

%% file: latex/abstract.tex
\begin{abstract}
    Pre-training models on Imagenet or other massive datasets of real images has led to major advances in computer vision, albeit accompanied with shortcomings related to curation cost, privacy, usage rights, and ethical issues. In this paper, for the first time, we study the transferability of pre-trained models based on synthetic data generated by graphics simulators to downstream tasks from very different domains.
    In using such synthetic data for pre-training, we find that downstream performance on different tasks are favored by different configurations of simulation parameters (\eg lighting, object pose, backgrounds, etc.), and that there is no one-size-fits-all solution. It is thus better to tailor synthetic pre-training data to a specific downstream task, for best performance. We introduce \ours, a unified model mapping downstream task representations to optimal simulation parameters to generate synthetic pre-training data for them. \ours learns this mapping by training to find the set of best parameters on a set of ``seen'' tasks. Once trained, it can then be used to predict best simulation parameters for novel ``unseen'' tasks in one shot, without requiring additional training. Given a budget in number of images per class, our extensive experiments with 20 diverse downstream tasks show \ours's task-adaptive pre-training data results in significantly better downstream performance than non-adaptively choosing simulation parameters on both seen and unseen tasks. It is even competitive with pre-training on real images from Imagenet.

\end{abstract}

%% file: latex/introduction.tex
\section{Introduction} \label{sec:intro}

\input{figures/fig1}

Using large-scale labeled (like ImageNet~\cite{deng2009imagenet}) or weakly-labeled (like JFT-300M~\cite{chollet2017xception,hinton2015distilling}, Instagram-3.5B~\cite{mahajan2018exploring}) datasets collected from the web has been the go-to approach for pre-training classifiers for \emph{downstream} tasks with a relative scarcity of labeled data. 
Prior works have demonstrated that as we move to bigger datasets for pre-training, downstream accuracy improves on average~\cite{sun2017revisiting,mahajan2018exploring}.
However, large-scale real image datasets bear the additional cost of curating labels, in addition to other concerns like privacy or copyright. Furthermore, large datasets like JFT-300M and Instagram-3.5B are not publicly available posing a bottleneck in reproducibility and fair comparison of algorithms.

Synthetic images generated via graphics engines provide an alternative quelling a substantial portion of these concerns. With 3D models and scenes, potentially infinite images can be generated by varying various scene or image-capture parameters. Although synthetic data has been used for transfer learning in various specialized tasks~\cite{su2015render, richter2016playing, tobin2017domain, anderson2021sim}, there has not been prior research dedicated to its transferability to a range of different recognition tasks from different domains (see \cref{fig:fig1}). In conducting this first of its kind (to the best of our knowledge) study, we first ask the question : in synthetic pretraining for different downstream classification tasks, does a one-size-fits-all solution (\ie, a universal pre-trained model for all tasks) work well?

With graphics engines, we can control various \emph{simulation} parameters (lighting, pose, materials, etc.). So, in an experiment, we introduced more variations successively from different parameters into a pretraining dataset of 100k synthetic images from 237 different classes (as many categories as are available in Three-D-World \cite{gan2020threedworld}). We pre-trained a ResNet-50 \cite{he2016deep} on these, and evaluated this backbone with linear probing on different downstream tasks. The results are in Table \ref{tab:motivation_results}. We see that some parameters like random object materials result in improved performance for some downstream tasks like SVHN and DTD, while hurting performance for other tasks like EuroSAT and Sketch. In general different pre-training data properties seem to favor different downstream tasks.

\input{tables/motivation_results}

To maximize the benefit of pre-training, different optimal simulation parameters can be found for each specific downstream task. Because of the combinatorially large set of different simulation parameter configurations, a brute force search is out of the question. However, this might still suggest that some, presumably expensive, learning process is needed for each downstream task for an optimal synthetic image set for pre-training. We show this is not the case. 

We introduce \ours, a unified model that maps a downstream task representation to optimal simulation parameters for pre-training data generation to maximize downstream accuracy. Using vector representations for a set of downstream tasks (in the form of Task2Vec \cite{achille2019task2vec}), we train \ours to find and thus learn a mapping to optimal parameters for each task from the set. Once trained on this set of ``seen'' tasks, \ours can also use Task2Vec representations of novel ``unseen'' tasks to predict simulation parameters that would be best for their pre-training datasets. This efficient one-shot prediction for novel tasks is of significant practical value, if developed as an end-user application that can automatically generate and provide pre-training data, given some downstream examples.

Our extensive experiments using $20$ downstream classification datasets show that on seen tasks, given a number of images per category, \ours's output parameters generate pre-training datasets that are much better for downstream performance than approaches like domain randomization~\cite{yue2019domain,anderson2021sim,khirodkar2019domain} that are not task-adaptive. Moreover, we show \ours also generalizes well to unseen tasks, maintaining an edge over non-adaptive approaches while being competitive with Imagenet pre-training.

In summary,
(i) We address a novel, and very practical, problem---how to optimally leverage synthetic data to task-adaptively pre-train deep learning models for transfer to diverse downstream tasks.
To the best of our knowledge, this is the first time such a problem is being addressed in transfer learning research.  
(ii) We propose \ours, a unified parametric model that learns to map Task2Vec representations of downstream tasks to simulation parameters for optimal pre-training. 
(iii) \ours can generalize to novel ``unseen'' tasks, not encountered during training, a feature of significant practical value as an application. 
(iv) We provide a thorough analysis of the behavior of downstream accuracy with different sizes of pre-training data (in number of classes, object-meshes or simply images) and with different downstream evaluation methods.

%% file: figures/fig1.tex
\begin{figure}[t]
    \centering
    \includegraphics[width=\linewidth,trim=0cm 0cm 0cm 0cm,clip]{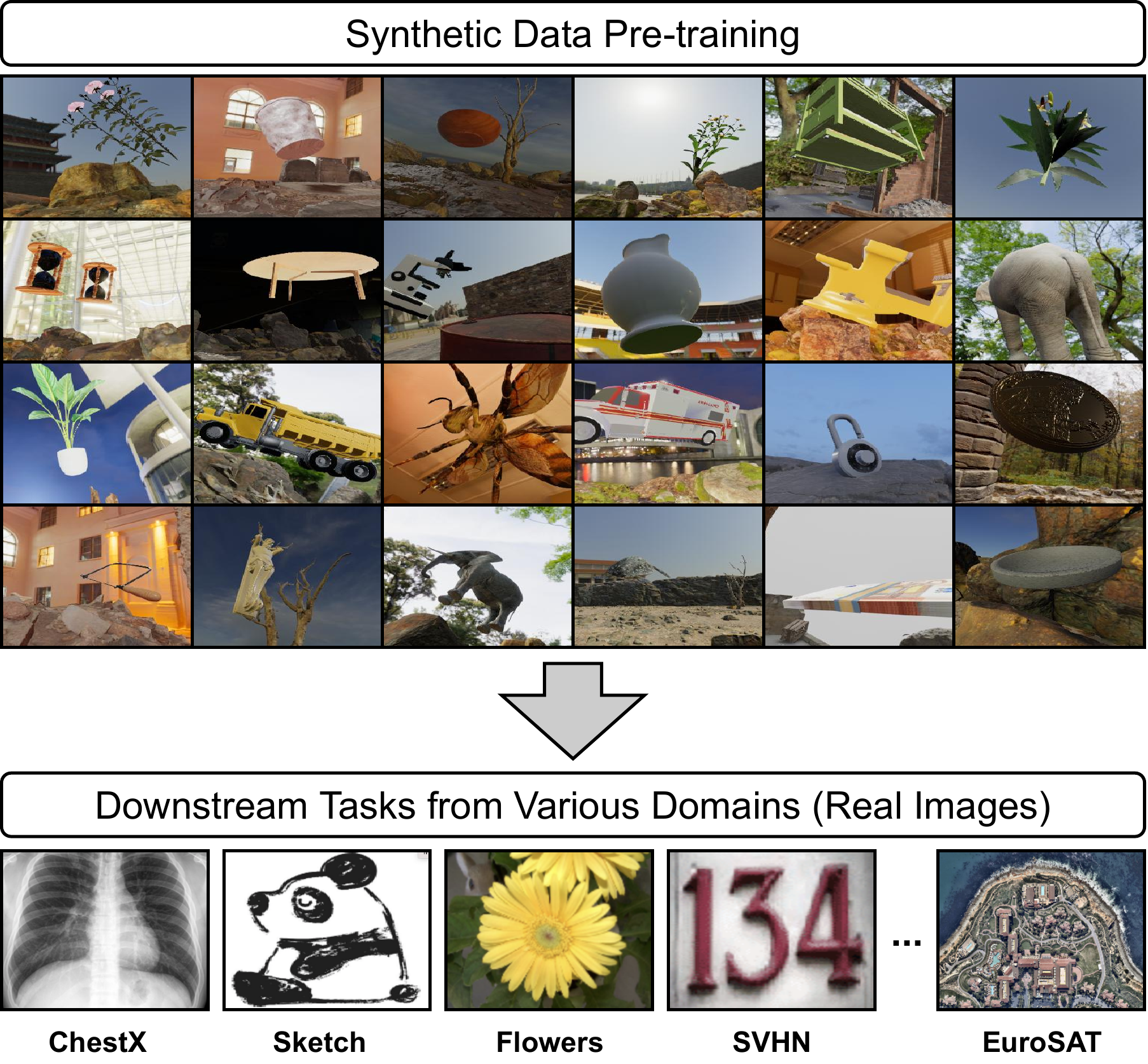}
   \vspace{-6mm}
    \caption{\small We explore how synthetic data can be effectively used for training models that can transfer to a wide range of downstream tasks from various domains. Is a universal pre-trained model for all downstream tasks, the best approach?}
    \label{fig:fig1} \vspace{-2mm}
\end{figure}

%% file: tables/motivation_results.tex
\begin{table}[t]
\centering
\scalebox{0.9}{
    \begin{tabular}{lcccc}
   \Xhline{2\arrayrulewidth}
    \multirow{2}[0]{*}{\thead{\textbf{Pretraining Data} \\ \textbf{Variations}}} & \multicolumn{4}{c}{\textbf{Downstream Accuracy}} \\
    \cline{2-5}
    & \textbf{EuroSAT} & \textbf{SVHN} & \textbf{Sketch} & \textbf{DTD} \\
    \midrule
    Pose  & 87.01 & 28.49 & 37.89 & 37.39 \\
    +Lighting & 88.57 & 32.36 & \textbf{38.81} & 40.32 \\
    +Blur & \textbf{90.20} & 35.58 & 35.53 & 37.66 \\
    +Materials & 84.54 & \textbf{44.84} & 30.81 & \textbf{38.51} \\
    +Background & 80.44 & 29.93 & 14.60 & 32.39 \\
   \Xhline{2\arrayrulewidth}
    \end{tabular}%
} \vspace{-2mm}
  \caption{\small Downstream task accuracies using linear probing with a Resnet-50 backbone pretrained on synthetic datasets with different varying parameters (successively added). We see different simulation parameters have different effects on downstream tasks.}
  \label{tab:motivation_results}%
  \vspace{-4mm}
\end{table}%

%% file: latex/related_work.tex
\vspace{-1mm}
\section{Related Work} \label{sec:related_work}

\noindent {\bf Training with Synthetic Data.} Methods that learn from synthetic data have been extensively studied since the early days of computer vision~\cite{nevatia1977description,little1988analysis}. In recent years,  many approaches that rely on synthetic data representations have been proposed for image classification~\cite{mikami2021scaling,gan2020threedworld}, object detection~\cite{peng2015learning,prakash2019structured}, semantic segmentation~\cite{wang2020differential,ros2016synthia}, action recognition~\cite{varol2021synthetic,roberto2017procedural}, visual reasoning~\cite{johnson2017clevr}, and embodied perception~\cite{savva2019habitat,kolve2017ai2,xia2018gibson}. While most of these rely on some graphics engines to generate synthetic images mimicking real ones, it has been observed that images seemingly consistent of noise can still be useful for representation learning~\cite{baradad2021learning}.
Unlike previous work, we focus on a different problem: how to build task-adaptive pre-trained models from synthetic data that can transfer to a wide range of downstream datasets from various domains.

\input{figures/controller-approach}

\vspace{1mm}
\noindent {\bf Synthetic to Real Transfer.} The majority of methods proposed to bridge the {\em reality gap} (between simulation and real data) are based on domain adaptation~\cite{csurka2017domain}. These include reconstruction-based techniques, using encoder-decoder models or GANs to improve the realism of synthetic data~\cite{richter2021enhancing,hoffman2018cycada,shrivastava2017learning},  discrepancy-based methods, designed to align features between the two domains~\cite{rozantsev2018beyond,zhang2019curriculum}, and adversarial approaches, which rely on a domain discriminator to encourage domain-independent feature learning~\cite{ren2018cross, ganin2016domain,tzeng2017adversarial}. Contrasting from these techniques, our work aims at building pre-trained models from synthetic data and does not assume the same label set for source and target domains. 
The most prevalent approach in a setting similar to ours, is domain randomization~\cite{tobin2017domain, prakash2019structured,yue2019domain,anderson2021sim,khirodkar2019domain}, which learns pre-trained models from datasets generated by randomly varying simulator parameters. In contrast, \ours \textit{learns} simulator parameters to generate synthetic datasets that maximize transfer learning performance.

\vspace{1mm}
\noindent {\bf Optimization of Simulator Parameters.} Recently, a few approaches have been proposed to learn synthetic data generation by optimizing simulator parameters~\cite{ruiz2018learning,kim2021drivegan,yang2020learning,behl2020autosimulate}. SPIRAL~\cite{ganin2018synthesizing}, AVO~\cite{louppe2019adversarial} and Attr. Desc.~\cite{yue2019domain} minimize the distance between distributions of simulated data and real data. Learning to Simulate~\cite{ruiz2018learning} optimizes simulator parameters using policy gradients that maximize validation accuracy for a specific task, while Auto-Sim~\cite{behl2020autosimulate} speeds up the search process using a differentiable approximation of the objective.
Meta-Sim~\cite{kar2019meta,devaranjan2020meta} learns to modify attributes obtained from probabilistic scene grammars for data generation.  These methods are specifically tailored to applications in autonomous driving, whereas our goal is to transfer synthetic data representations to a wide range of downstream tasks. Notably, our proposed approach is significantly different from previous methods, as it maps task representations to simulation parameters through a unified parametric model, enabling one-shot synthetic data generation, even for unseen tasks, without requiring expensive training.

\vspace{1mm}
\noindent {\bf Conditional Computation.} Albeit not apparent, our method is also related to dynamic neural network models that adaptively change computation depending on the input~\cite{han2021dynamic}. These methods have been effectively used to skip computation in deep neural networks conditioned on the input~\cite{wu2018blockdrop,wang2018skipnet,veit2018convolutional}, perform adaptive fine-tuning~\cite{guo2019spottune}, and dynamically allocate computation across frames for efficient video analysis~\cite{wu2019adaframe,meng2020ar}. In particular, Adashare~\cite{sun2019adashare} learns different computational pathways for each task within a single multi-task network model, with the goal of improving efficiency and minimizing negative interference in multi-task learning. Analogously, our approach learns different {\em data simulation pathways} (by adaptively deciding which rendering parameters to use) for each task, using a single parametric model, with the goal of generating task-specific pre-training data.

%% file: figures/controller-approach.tex
\begin{figure*}[h]
    \centering
    \includegraphics[width=\linewidth,trim=0cm 0cm 0cm 0cm,clip]{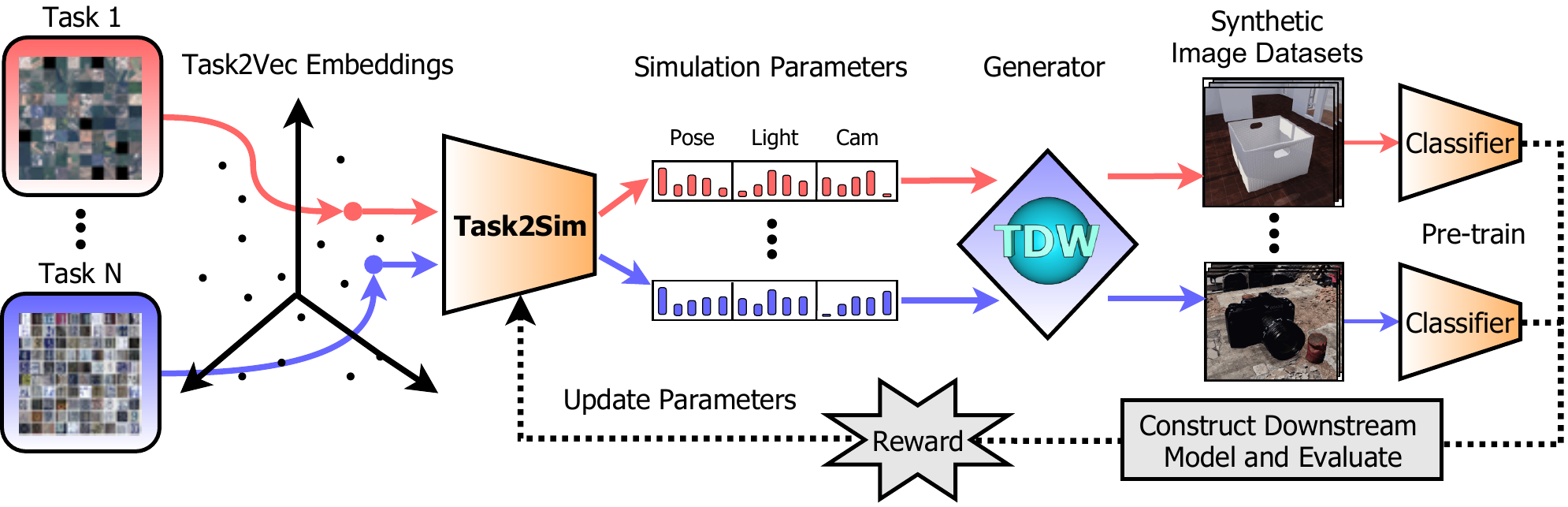}
    \vspace{-6mm}
    \caption{\small \textbf{Illustration of our proposed approach}. Given a batch of tasks represented by Task2Vec representations, our approach (\ours) aims to map these representations to optimal simulation parameters for generating a dataset of synthetic images. The downstream classifier’s accuracy for the set of tasks is then used as a reward to update \ours's parameters. Once trained, \ours can be used not only for “seen” tasks but also can be used in one-shot to generate simulation parameters for novel “unseen” tasks.}
    \label{fig:controller-approach} \vspace{-2mm}
\end{figure*}

%% file: latex/method.tex
\section{Proposed Approach} \label{sec:method}

Our goal is to create a unified model that maps task representations (e.g., obtained using task2vec~\cite{achille2019task2vec}) to simulation parameters, which are in turn used to render synthetic pre-training datasets for not only tasks that are seen during training, but also novel tasks.
This is a challenging problem, as the number of possible simulation parameter configurations is combinatorially large, making a brute-force approach infeasible when the number of parameters grows. 

\subsection{Overview} 

\cref{fig:controller-approach} shows an overview of our approach. During training, a batch of ``seen'' tasks is provided as input. Their task2vec vector representations are fed as input to \ours, which is a parametric model (shared across all tasks) mapping these downstream task2vecs to simulation parameters, such as lighting direction, amount of blur, background variability, etc.  These parameters are then used by a data generator (in our implementation, built using the Three-D-World platform~\cite{gan2020threedworld}) to generate a dataset of synthetic images. A classifier model then gets pre-trained on these synthetic images, and the backbone is subsequently used for evaluation on specific downstream task. The classifier's accuracy on this task is used as a reward to update \ours's parameters. 
Once trained, \ours can also be used to efficiently predict simulation parameters in {\em one-shot} for ``unseen'' tasks that it has not encountered during training.

\subsection{\ours Model}

Let us denote \ours's parameters with $\theta$. Given the task2vec representation of a downstream task $\bs{x} \in \mc{X}$ as input, \ours outputs simulation parameters $a \in \Omega$. The model consists of $M$ output heads, one for each simulation parameter. In the following discussion, just as in our experiments, each simulation parameter is discretized to a few levels to limit the space of possible outputs. Each head outputs a categorical distribution $\pi_i(\bs{x}, \theta) \in \Delta^{k_i}$, where $k_i$ is the number of discrete values for parameter $i \in [M]$, and $\Delta^{k_i}$, a standard $k_i$-simplex. The set of argmax outputs $\nu(\bs{x}, \theta) = \{\nu_i | \nu_i = \argmax_{j \in [k_i]} \pi_{i, j} ~\forall i \in [M]\}$ is the set of simulation parameter values used for synthetic data generation. Subsequently, we drop annotating the dependence of $\pi$ and $\nu$ on $\theta$ and $\bs{x}$ when clear.

\subsection{\ours Training}

Since Task2Sim aims to maximize downstream accuracy after pre-training, we use this accuracy as the reward in our training optimization\footnote{Note that our rewards depend only on the task2vec input and the output action and do not involve any states, and thus our problem can be considered similar to a stateless-RL or contextual bandits problem \cite{langford2007epoch}.}.
Note that this downstream accuracy is a non-differentiable function of the output simulation parameters (assuming any simulation engine can be used as a black box) and hence direct gradient-based optimization cannot be used to train \ours. Instead, we use REINFORCE~\cite{williams1992simple}, to approximate gradients of downstream task performance with respect to model parameters $\theta$. 

\ours's outputs represent a distribution over ``actions'' corresponding to different values of the set of $M$ simulation parameters. $P(a) = \prod_{i \in [M]} \pi_i(a_i)$ is the probability of picking action $a = [a_i]_{i \in [M]}$, under policy $\pi = [\pi_i]_{i \in [M]}$. Remember that the output $\pi$ is a function of the parameters $\theta$ and the task representation $\bs{x}$. To train the model, we maximize the expected reward under its policy, defined as
\begin{align}
    R = \E_{a \in \Omega}[R(a)] = \sum_{a \in \Omega} P(a) R(a)
\end{align}
where $\Omega$ is the space of all outputs $a$ and $R(a)$ is the reward when parameter values corresponding to action $a$ are chosen. Since reward is the downstream accuracy, $R(a) \in [0, 100]$.  
Using the REINFORCE rule, we have
\begin{align}
    \nabla_{\theta} R 
    &= \E_{a \in \Omega} \left[ (\nabla_{\theta} \log P(a)) R(a) \right] \\
    &= \E_{a \in \Omega} \left[ \left(\sum_{i \in [M]} \nabla_{\theta} \log \pi_i(a_i) \right) R(a) \right]
\end{align}
where the 2nd step comes from linearity of the derivative. In practice, we use a point estimate of the above expectation at a sample $a \sim (\pi + \epsilon)$ ($\epsilon$ being some exploration noise added to the Task2Sim output distribution) with a self-critical baseline following \cite{rennie2017self}:
\begin{align} \label{eq:grad-pt-est}
    \nabla_{\theta} R \approx \left(\sum_{i \in [M]} \nabla_{\theta} \log \pi_i(a_i) \right) \left( R(a) - R(\nu) \right) 
\end{align}
where, as a reminder $\nu$ is the set of the distribution argmax parameter values from the \name{} model heads.

A pseudo-code of our approach is shown in \cref{alg:train}.  Specifically, we update the model parameters $\theta$ using minibatches of tasks sampled from a set of ``seen'' tasks. Similar to \cite{oh2018self}, we also employ self-imitation learning biased towards actions found to have better rewards. This is done by keeping track of the best action encountered in the learning process and using it for additional updates to the model, besides the ones in \cref{ln:update} of \cref{alg:train}. 
Furthermore, we use the test accuracy of a 5-nearest neighbors classifier operating on features generated by the pretrained backbone as a proxy for downstream task performance since it is computationally much faster than other common evaluation criteria used in transfer learning, e.g., linear probing or full-network finetuning. Our experiments demonstrate that this proxy evaluation measure indeed correlates with, and thus, helps in final downstream performance with linear probing or full-network finetuning.

\begin{algorithm}
\DontPrintSemicolon
 \textbf{Input:} Set of $N$ ``seen'' downstream tasks represented by task2vecs $\mc{T} = \{\bs{x}_i | i \in [N]\}$. \\
 Given initial Task2Sim parameters $\theta_0$ and initial noise level $\epsilon_0$\\
 Initialize $a_{max}^{(i)} | i \in [N]$ the maximum reward action for each seen task \\
 \For{$t \in [T]$}{
 Set noise level $\epsilon = \frac{\epsilon_0}{t} $ \\
 Sample minibatch $\tau$ of size $n$ from $\mc{T}$  \\
 Get \ours output distributions $\pi^{(i)} | i \in [n]$ \\
 Sample outputs $a^{(i)} \sim \pi^{(i)} + \epsilon$ \\
 Get Rewards $R(a^{(i)})$ by generating a synthetic dataset with parameters $a^{(i)}$, pre-training a backbone on it, and getting the 5-NN downstream accuracy using this backbone \\
 Update $a_{max}^{(i)}$ if $R(a^{(i)}) > R(a_{max}^{(i)})$ \\
 Get point estimates of reward gradients $dr^{(i)}$ for each task in minibatch using \cref{eq:grad-pt-est} \\
 $\theta_{t,0} \leftarrow \theta_{t-1} + \frac{\sum_{i \in [n]} dr^{(i)}}{n}$ \label{ln:update} \\
 \For{$j \in [T_{si}]$}{ 
    \tcp{Self Imitation}
    Get reward gradient estimates $dr_{si}^{(i)}$ from \cref{eq:grad-pt-est} for $a \leftarrow a_{max}^{(i)}$ \\
    $\theta_{t, j}  \leftarrow \theta_{t, j-1} + \frac{\sum_{i \in [n]} dr_{si}^{(i)}}{n}$
 }
 $\theta_{t} \leftarrow \theta_{t, T_{si}}$
 }
 \textbf{Output}: Trained model with parameters $\theta_T$. 
 \caption{Training Task2Sim}
 \label{alg:train}  
\end{algorithm}

%% file: latex/experiments.tex
\section{Experiments} \label{sec:expts}

\subsection{Details} \label{subsec:expts_details}

\noindent \textbf{Downstream Tasks.} We use a set of 20 classification tasks with
12 tasks from \cite{islam2021broad} as the set of ``seen'' tasks for our model and a separate set of 8 tasks as the ``unseen'' set.
All our tasks can be broadly categorized into the following 6 classes (S:seen, U:unseen): 
\begin{itemize}
\vspace{-0.5em}
\itemsep-0.2em
\item Natural Images: CropDisease(S)\cite{mohanty2016cropdisease}, Flowers102(S)\cite{nilsback2008automatedflowers102}, DeepWeeds(S)\cite{DeepWeeds2019}, CUB(U)\cite{WahCUB_200_2011}
\item Aerial Images: EuroSAT(S)\cite{helber2019eurosat}, Resisc45(S)\cite{cheng2017remoteresisc45}, AID(U)\cite{xia2017aid}, CactusAerial(U)\cite{lopez2019columnarCactusAerial}
\item Symbolic Images: SVHN(S)\cite{netzer2011readingsvhn}, Omniglot(S)\cite{lake2015humanomniglot}, USPS(U)\cite{hull1994databaseUSPS}
\item Medical Images: ISIC(S)\cite{codella2019skinisic}, ChestX(S)\cite{wang2017chestx}, ChestXPneumonia(U)\cite{kermany2018identifyingChestXP}
\item Illustrative Images: Kaokore(S)\cite{tian2020kaokore}, Sketch(S)\cite{wang2019learningsketch}, Pacs-C(U), Pacs-S(U)\cite{li2017deeperPACS}
\item Texture Images: DTD(S)\cite{cimpoi2014DTD}, FMD(U)\cite{zhang2019poissonFMD}
\end{itemize}
\vspace{-0.3em}

\noindent \textbf{Task2Sim Details.} We used a Resnet-18 probe network to generate 9600-dimensional Task2Vec representations of downstream tasks. The Task2Sim model is a multi-layer perceptron with 2 hidden layers, having ReLU activations. The model shares its first two layers for all $M$ heads, and branches after that. It is trained for 1000 epochs on seen tasks, with a batch-size 4 and 5 self-imitation steps (\ie $n=4, T_{si} = 5$ and $T = 1000$). We used a Resnet-50 model for pre-training and downstream evaluation for \ours's rewards. Details of all datasets, pre-training and evaluation procedures are included in Appendix \ref{sec:train_details}.

\vspace{1mm}
\noindent \textbf{Synthetic Data Generation.} We use Three-D-World (TDW) \cite{gan2020threedworld} for synthetic image generation. The platform provides 2322 different object models from 237 different classes, 57 of which overlap with Imagenet. Using TDW, we generate synthetic images of single objects from the aforementioned set (see Figure~\ref{fig:fig1} for examples). 

In this paper, we experiment with a parameterization of the pretraining dataset where $M = 8$ and $k_i = 2 ~\forall~ i \in [M]$ (using terminology from Section \ref{sec:method}). The 8 parameters are: 
\begin{itemize}
\vspace{-0.3em}
\itemsep-0.2em
    \item Object Rotation : If $1$, multiple poses of an object are shown in the dataset, else, an object appears in a canonical pose in each image.
    \item Object Distance (from camera) : If $1$, object distance from the camera is varied randomly within a certain range, else, it is kept fixed.
    \item Lighting Intensity : If $1$, intensity of the main lighting source (sun-like point light source at a distance) is varied, else it is fixed.
    \item Lighting Color : If $1$, RGB color of the main lighting source is varied, else it is fixed.
    \item Lighting Direction : If $1$, the direction of the main light source is varied, else it is a constant.
    \item Focus Blur : If $1$, camera focus point and aperture are randomly perturbed to induce blurriness in the image, else, all image contents are always in focus.
    \item Background : If $1$, background of the object changes in each image, else it is held fixed.
    \item Materials : If $1$, in each image, each component of an object is given a random material out of 140 different materials, else objects have their default materials. 
\end{itemize}
\vspace{-0.5em}
Hence in our experiments, for each of the above 8 parameters, \ours decided whether or not different variations of it, would exhibit in the dataset. For speed of dataset generation while training \ours, we used a subset of 780 objects with simple meshes, from 100 different categories and generated 400 images per category for pre-training.

\subsection{\name~Results}

\noindent \textbf{Baselines.} We compared Task2Sim's downstream performance with the following baselines (pre-training datasets):
(1) Random : For each downstream dataset, chooses a random 8-length bit string as the set of simulation parameters. 
(2) Domain Randomization : Uses a 1 in each simulation parameter, thus using all variations from simulation in each image. 
\interfootnotelinepenalty=10000
(3) Imagenet : Uses a subset of Imagenet with equal number of classes and images as other baselines\footnote{ We also compared pre-training using Imagenet with 1K classes and an equal number of images, but this was poorer on average in downstream performance than the subset with fewer classes. \cref{tab:main_seen,,tab:main_unseen} and \cref{fig:main_all_seen,,fig:main_all_unseen} do not include it for succinctness.}.
(4)~Scratch : Does not involve any pre-training of the classifier's feature extractor, training a randomly initialized classifier, with only downstream task data.

\input{tables/main_seen}

\input{tables/main_unseen}

\input{figures/main_all_seen}
\input{figures/main_all_unseen}

\vspace{1mm}
\noindent \textbf{Performance on Seen Tasks.} \cref{tab:main_seen} shows accuracies averaged over 12 seen downstream tasks for \ours and all baselines using different evaluation methods for a Resnet-50 backbone. For the last two columns, we included all objects of TDW from 237 categories, and kept the number of images at roughly 400 per class, resulting in about 100k images total, regenerating a new dataset with the simulation parameters corresponding to the different synthetic image generation methods. On average, over the 12 seen tasks, simulation parameters that \ours finds are better than Domain Randomization and Random selection and are competitive with Imagenet pre-training, both for the subset of classes that Task2Sim is trained using, and when a larger set of classes is used. \cref{fig:main_all_seen} shows accuracies for the 12 seen datasets for different methods, on the 237 category 100k image pre-training set.

\vspace{1mm}
\noindent \textbf{Performance on Unseen Tasks.} \cref{tab:main_unseen} shows average downstream accuracy over 8 unseen datasets, of a Resnet-50 pretrained on different datasets. 
We see that Task2Sim generalizes well, and is still better than Domain Randomization and Random simulation parameter selection. Moreover, it is marginally better on average than Imagenet pretraining for these tasks. \cref{fig:main_all_unseen} shows the accuracies from the last column of \cref{tab:main_unseen} over the 8 individual unseen tasks.

\subsection{Analysis} \label{subsec:analysis}
\noindent \textbf{Task2Sim Outputs.}
\cref{fig:t2s_outputs} shows the output distribution from the trained Task2Sim model for different seen and unseen tasks. Each output shows the probability assigned by the model to the output 1 in that particular simulation parameter. From the outputs, we see the model determines that in general for the set of tasks considered, it is better to see a single pose of objects rather than multiple poses, and that it is better to have scene lighting intensity variations in different images than have lighting of constant intensity in all images. In general, adding material variability was determined to be worse for most datasets, except for SVHN. Comparing predictions for seen vs unseen tasks, we see that Task2Sim does its best to generalize to unseen tasks by relating them to the seen tasks. 
For \eg, outputs for ChestXPneumonia are similar to ChestX, while outputs of CactusAerial are similar to those of EuroSAT, both being aerial/satellite image datasets. A similar trend is also seen in PacsS and Sketch both of which contain hand-sketches, and for CUB and CropDisease, both natural image datasets.

\input{figures/t2s_outputs}

Another inspection shows Task2Sim makes decisions that are quite logical for certain tasks. For instance, Task2Sim turns off the ``Light Color'' parameter for CUB. Here, color plays a major role in distinguishing different birds, thus needing a classifier representation that should not be invariant to color changes. Indeed, from \cref{fig:nn}, we see that the neighbors of Task2Sim are of similar colors.

\input{figures/diff_num_cls}

\input{figures/diff_num_objs}

\input{figures/diff_num_imgs}

\input{figures/nn}

\vspace{1mm}
\noindent \textbf{Effect of Number of Pretraining Classes.}
In \cref{fig:diff_num_cls}, we plot the average accuracy with full network finetuning on the 12 seen downstream tasks. On the x-axis, we vary the number of classes used for pre-training, with 1000 images per class on average (200 classes=200k images). We see all pre-training methods improve with more classes (and correspondingly more images) at about similar rates. Task2Sim stays better than Domain Randomization and competitive with (about 2\% shy of) pre-training with an equivalent subset (in number of classes and images) of Imagenet.

\vspace{1mm}
\noindent \textbf{Effect of Number of Different Objects per Class.} In TDW, we have 2322 object meshes from 237 different categories. In \cref{fig:diff_num_objs}, we vary the number of object meshes used per category. The point right-most on the x-axis has 200k images with all objects used, and moving to the left, the number of images reduces proportionately as a fraction of these objects are used (the number of categories being the same). We find that with increasing number of different objects used for each category, Domain Randomization improves downstream performance at a slightly higher rate than our proposed Task2Sim.

\vspace{1mm}
\noindent \textbf{Effect of Number of Pretraining Images.} In \cref{fig:diff_num_imgs}, we show average downstream task accuracy, for the 12 seen tasks, with different number of images used for pretraining. All methods, except Imagenet-1K and Scratch, use 237 image categories, with synthetic datasets using all available object models. Imagenet-237 is a subset of Imagenet-1K containing 237 categories that were randomly picked.
We see Task2Sim is highly effective in the regime where fewer images are available for pre-training, and is even slightly better than pre-training with Imagenet at 50k images. It maintains its advantage over non-adaptive pretraining up to a significant extrapolation of 500k images, having only trained using smaller datasets (of 100 classes and 40k images). At 1M images, it is still competitive with Imagenet pre-training and is much better than training from scratch. 

We also observe that all methods improve when more pre-training images are available, although the rate of improvement decreases as we move along positive X-direction.
Initially, Domain Randomization improves at a higher rate than Task2Sim and at 1M pretraining images, matches its performance. This is likely because at a higher number of images, even when there are all variations possible from simulation in each image (corresponding to Domain Randomization), the deep feature extractor grows robust to the variations which may not add any value to the representation for specific downstream tasks. 

Our hypothesis is that at a fixed number of categories there may exist some point in number of pre-training images when the above robustness can be good enough to match our Task2Sim's downstream performance. With a 237-category limit from TDW and using the set of variations from our 8 chosen parameters, 1M images seems to be this point. However as the number of classes increases, this point shifts towards higher number of images. As evidence, consider \cref{fig:diff_num_cls}, where we see that as more classes of objects are added with more data, different methods improve at similar rates. Moving further along positive X, if this holds with more classes, Task2Sim maintains its edge over Domain Randomization even at higher numbers of images. This suggests a non-adaptive pre-training method like Domain Randomization has potential to be as effective on average as Task2Sim, but only at the cost of more pre-training images. However, this cost would keep increasing as pre-training data encompasses more object categories, and would be unknown without experimentation. 

For additional results and discussions, we refer readers to the Appendix.

%% file: tables/main_seen.tex
\begin{table*}[h!]
\centering
\scalebox{1.0}{
  
    \begin{tabular}{l|ccc|cc}
   \Xhline{2\arrayrulewidth}
          & \multicolumn{5}{c}{\textbf{Average Downstream Accuracy --- \texttt{Seen Tasks}}} \\
          \cline{2-6}
          & \multicolumn{3}{c|}{\textbf{100 classes / 40k images}} & \multicolumn{2}{c}{\textbf{237 classes / 100k images}} \\
          \cline{2-6}
    \multicolumn{1}{c|}{\textbf{Pretraining Dataset}} & \thead{\textbf{5NN}} & \thead{\textbf{Linear Probing} } & \thead{\textbf{Finetuning}} & \thead{\textbf{Linear Probing}} & \thead{\textbf{Finetuning}} \\
   \Xhline{2\arrayrulewidth}
    Scratch &  -  &  -  & 64.85 & -  & 64.85 \\
    Random & 25.30 & 54.06 & 70.77 & 55.14 & 72.18 \\
    Domain Randomization & 19.42 & 35.31 & 62.96 & 45.31 & 68.51 \\
    Imagenet$^{*}$ & \underline{28.91} & \textbf{63.12} & \underline{74.26} & \textbf{68.44} & \textbf{77.61} \\
    \texttt{\textbf{\ours}} & \textbf{30.46} & \underline{62.70} & \textbf{75.34} & \underline{62.71} & \underline{76.87} \\
   \Xhline{2\arrayrulewidth}
    \end{tabular}%
} \vspace{-2mm}
  \caption{Comparing the downstream accuracy on seen tasks for the Task2Sim chosen pretraining dataset and other baselines. Simulation parameters found on seen tasks by \ours generates synthetic pretraining data that is better for downstream tasks than other approaches like using Random simulation parameters or Domain Randomization. Pre-training with \ours's data is also competitive with pre-training on images from Imagenet. $^{*}$Imagenet has been subsampled to the same number of classes and images as indicated at the top of the column. boldface=highest, underline=$2^{nd}$ highest in column.}
  \label{tab:main_seen}%
  \vspace{-2mm}
\end{table*}%

%% file: tables/main_unseen.tex
\begin{table*}[h!]
\centering
\scalebox{1}{
  
    \begin{tabular}{l|ccc|cc}
    \Xhline{2\arrayrulewidth}
          & \multicolumn{5}{c}{\textbf{Average Downstream Accuracy --- \texttt{Unseen Tasks}}} \\
    \cline{2-6}
    & \multicolumn{3}{c|}{\textbf{100 classes / 40k images}} & \multicolumn{2}{c}{\textbf{237 classes / 100k images}} \\
    \cline{2-6}
    \multicolumn{1}{c|}{\textbf{Pretraining Dataset}} & \thead{\textbf{5NN} } & \thead{\textbf{Linear Probing}} & \thead{\textbf{Finetuning}} & \thead{\textbf{Linear Probing}} & \thead{\textbf{Finetuning}} \\
    \Xhline{2\arrayrulewidth}
    Scratch &   -    &  -     & 76.86 & -      & 76.86 \\
    Random & 51.80 & 74.68 & 83.97 & 74.11 & 83.49 \\
    Domain Randomization & 45.06 & 56.96 & 72.64 & 69.12 & 78.15 \\
    Imagenet$^*$ & \textbf{54.12} & \underline{75.47} & \underline{84.78} & \underline{81.33} & \underline{87.84} \\
    \texttt{\textbf{\ours}} & \underline{53.06} & \textbf{79.25} & \textbf{87.05} & \textbf{82.05} & \textbf{88.77} \\
\Xhline{2\arrayrulewidth}
    \end{tabular}%
} \vspace{-2mm}
  \caption{Comparing the downstream accuracy on unseen tasks for the \ours chosen pretraining dataset and other baselines. \ours also generalizes well to ``unseen'' tasks, not encountered during training, maintaining an edge over other synthetic data, while still being competitive with Imagenet. $^{*}$Imagenet subsampled as in \cref{tab:main_seen}. boldface=highest, underline=$2^{nd}$ highest in column.}
  \label{tab:main_unseen}%
  \vspace{-3mm}
\end{table*}%

%% file: figures/main_all_seen.tex
\begin{figure}[h]
    \centering
    \includegraphics[width=0.9\linewidth,trim=0cm 0cm 0cm 0cm,clip]{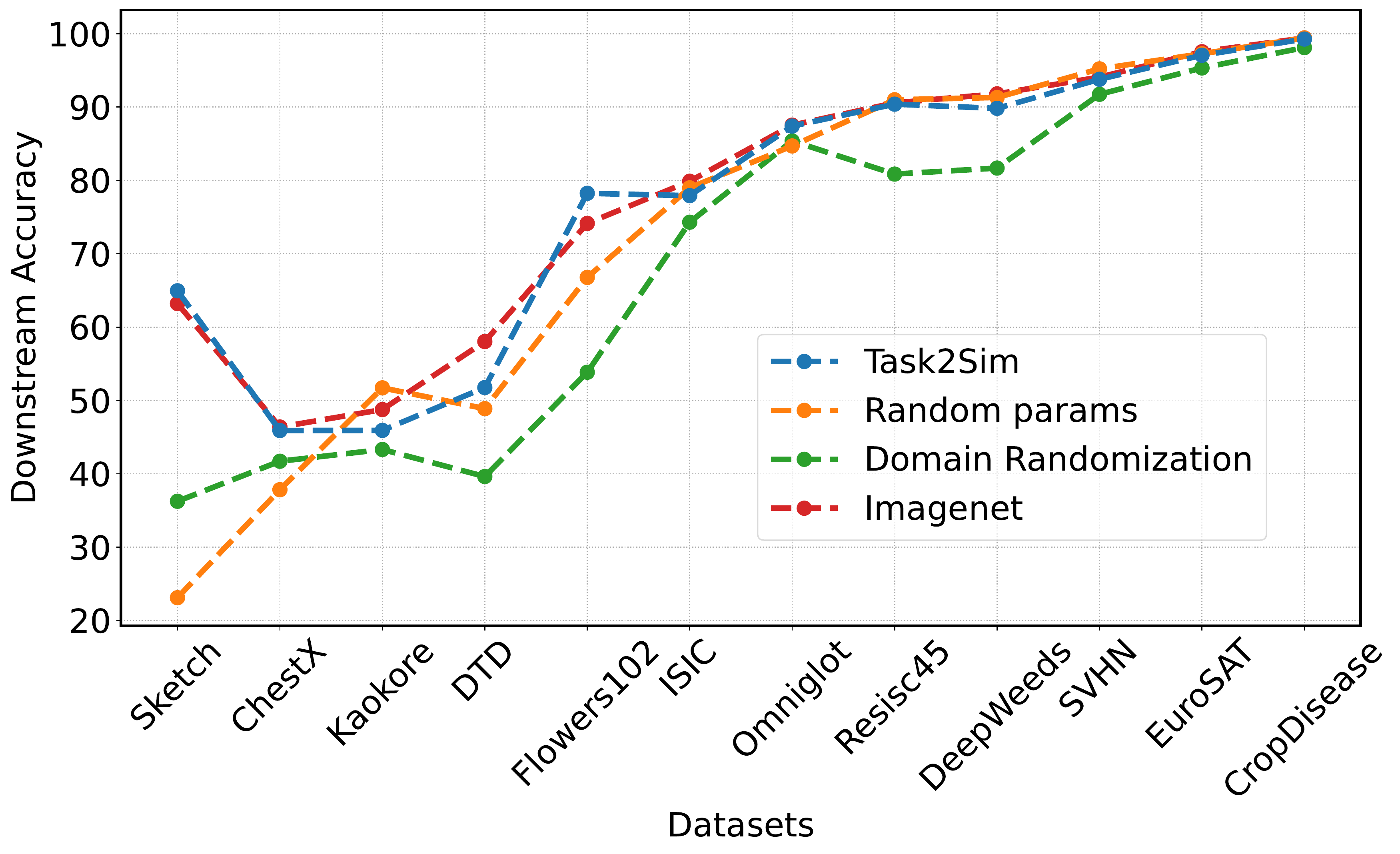}
    \vspace{-2mm}
    \caption{Performance of Task2Sim vs baselines on 12 seen tasks for 237 class / 100k image pre-training datasets evaluated with full-network finetuning. Best viewed in color.}
    \label{fig:main_all_seen}
   \vspace{-3mm}
\end{figure}

%% file: figures/main_all_unseen.tex
\begin{figure}[h]
    \centering
    \includegraphics[width=0.9\linewidth,trim=0cm 0cm 0cm 0cm,clip]{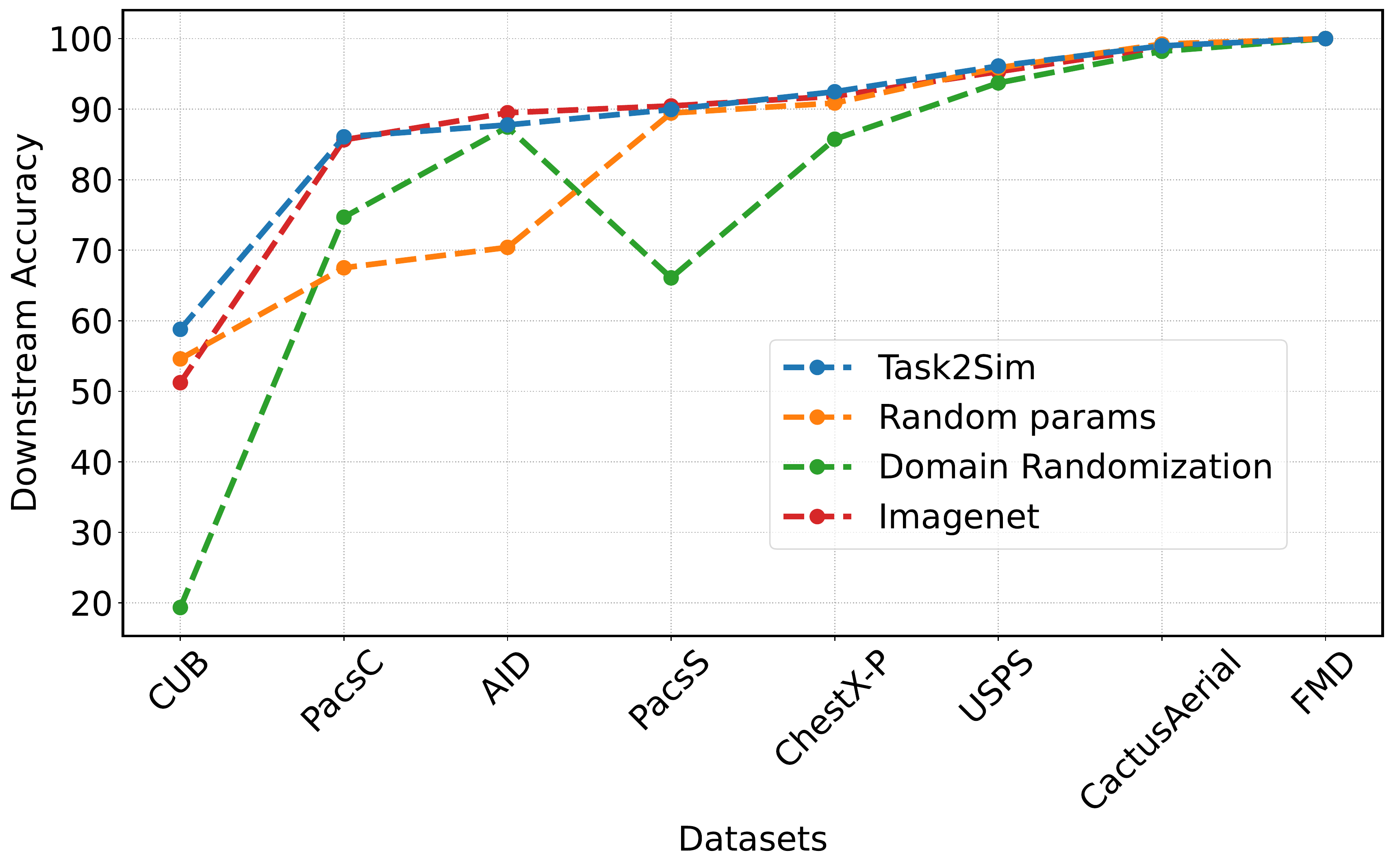}
    \vspace{-2mm}
    \caption{Performance of Task2Sim vs baselines on 8 unseen tasks for 237 class / 100k image pre-training datasets evaluated with full-network finetuning. Best viewed in color.}
    \label{fig:main_all_unseen}
    \vspace{-3mm}
\end{figure}

%% file: figures/t2s_outputs.tex
\begin{figure}[b]
    \centering
  \vspace{-4mm}
    \includegraphics[width=\linewidth,trim=0cm 0cm 0cm 0cm,clip]{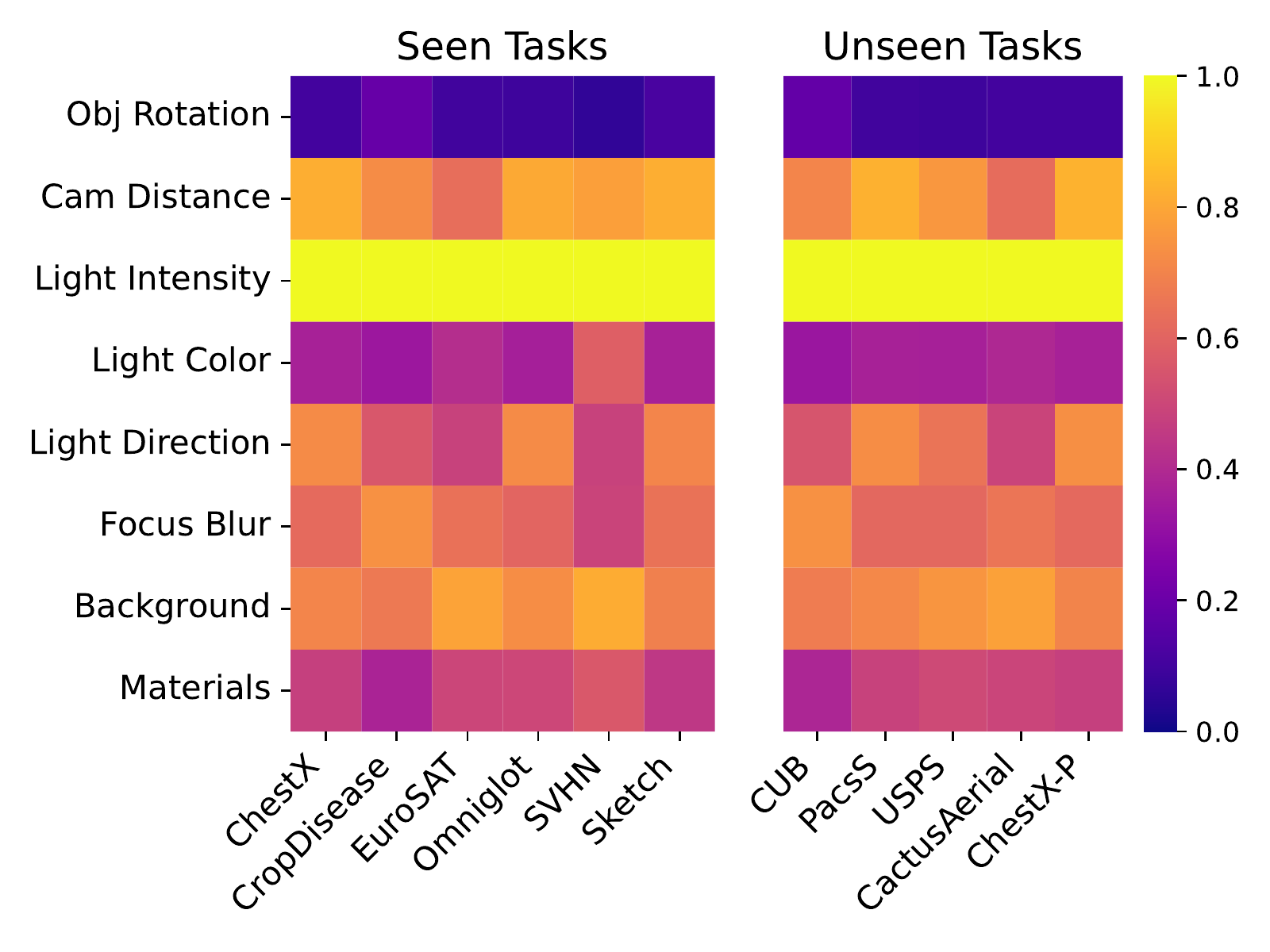}
    \caption{Task2Sim outputs for different seen and unseen tasks. Values shown are predicted probability of value 1 in the specific simulation parameters. Best viewed in color.}
    \label{fig:t2s_outputs}
\end{figure}

%% file: figures/diff_num_cls.tex
\begin{figure}[h]
    \centering
    \includegraphics[width=\linewidth,trim=0cm 0cm 0cm 0cm,clip]{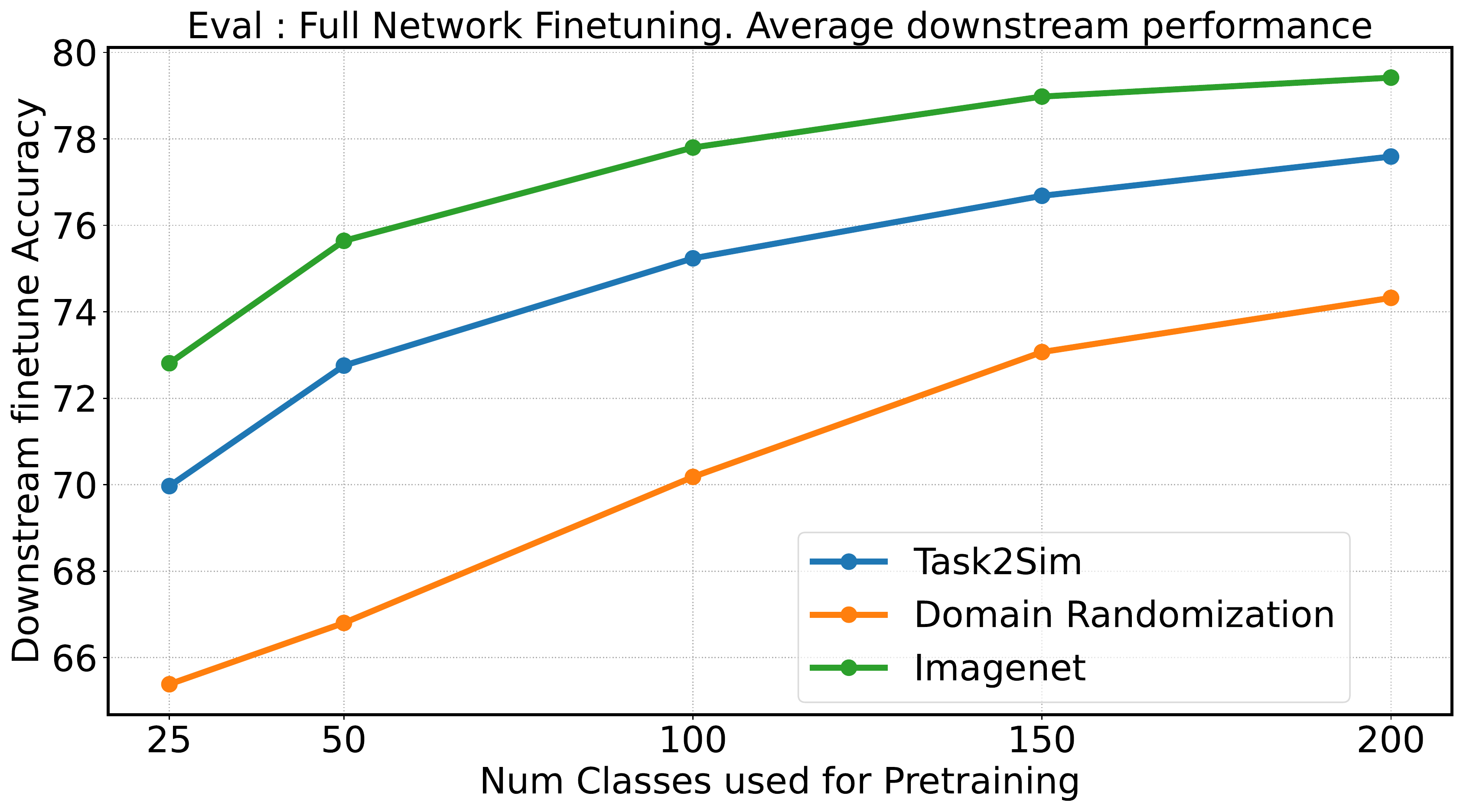}
  \vspace{-4mm}
    \caption{Avg performance over 12 seen tasks at different number of classes for pre-training. All methods improve performance at similar rates with the addition of more classes.}
    \label{fig:diff_num_cls} \vspace{-2mm}
\end{figure}

%% file: figures/diff_num_objs.tex
\begin{figure}[h]
    \centering
    \includegraphics[width=\linewidth,trim=0cm 0cm 0cm 0cm,clip]{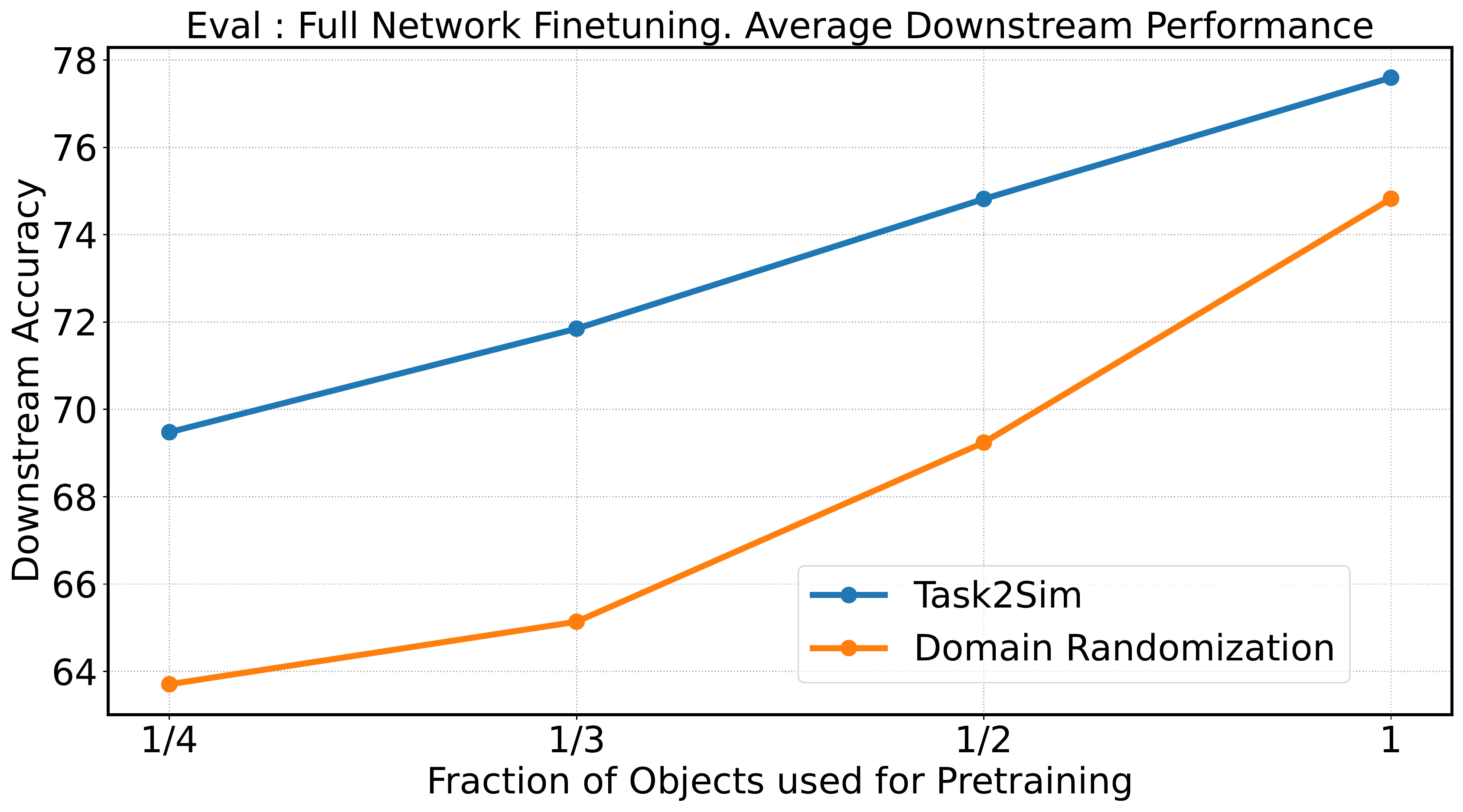}
    \vspace{-4mm}
    \caption{Avg performance over 12 seen tasks at different number of object meshes used per category for generating synthetic pretraining data. Both methods of synthetic data generation improve performance with addition of more objects with Domain Randomization improving at a slightly higher rate.}
    \label{fig:diff_num_objs} \vspace{-2mm}
\end{figure}

%% file: figures/diff_num_imgs.tex
\begin{figure}[]
    \centering
    \centering
    \includegraphics[width=\linewidth]{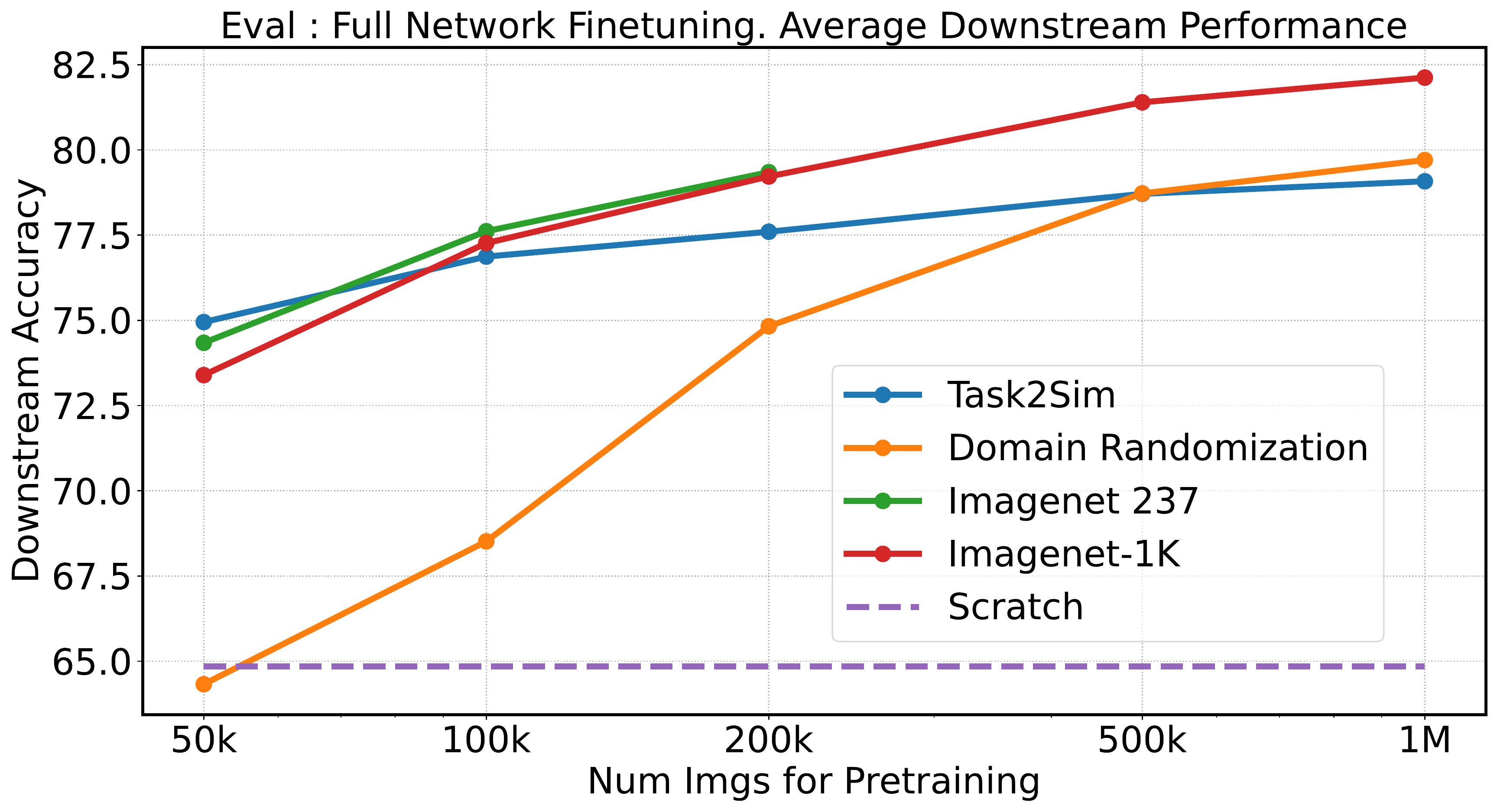}
    \vspace{-4mm}
    \caption{Task2Sim performance (avg over 12 seen tasks) vs other methods at different number of images for pretraining. Task2Sim is highly effective at fewer images. Increasing the number of images improves performance for all methods, reaching a saturation at a high enough number. See \cref{subsec:analysis} for more discussion.}
    \label{fig:diff_num_imgs}
    \vspace{-2mm}
\end{figure}

%% file: figures/nn.tex
\begin{figure*}[t]
    \centering
    \includegraphics[width=\linewidth,trim=0cm 0cm 0cm 0cm,clip]{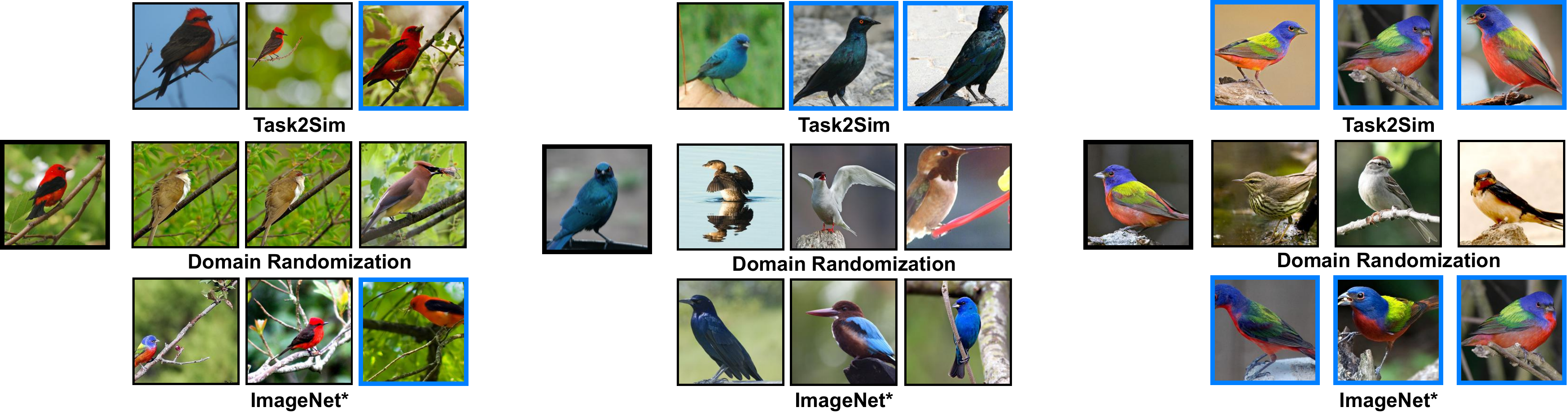}
    \vspace{-5mm}
    \caption{3 nearest neighbors of three test examples from the CUB dataset based on different pretrained feature representations (top: Task2Sim, middle: domain randomization, bottom: ImageNet*). Neighbors with a blue box share the same class with the anchor image on the left. For Task2Sim, similar to Imagenet, the neighbors are of similar color which suggests that the pretrained representation captures color similarity which can be crucial for identifying different bird species. Best viewed in color.}
    \label{fig:nn} \vspace{-3mm}
\end{figure*}

%% file: latex/conclusion.tex
\vspace{-1mm}
\section{Conclusion} \label{sec:conclusion}
\vspace{-1mm}

We saw the approach that is optimal for downstream performance in using synthetic data for pre-training is to specifically adapt the synthetic data to different downstream tasks. In this paper, we parameterized our synthetic data via different simulation parameters from graphics engines, and introduced \ours, which learns to map downstream task representations to optimal simulation parameters for synthetic pretraining data for the task. We showed \ours can be trained on a set of ``seen'' tasks and can then generalize to novel ``unseen'' tasks predicting parameters for them in one-shot, making it highly practical for synthetic pre-training data generation. While a large portion of contemporary representation learning research focuses on self-supervision to avoid using labels, we hope our demonstration with Task2Sim motivates further research in using simulated data from graphics engines for this purpose, with focus on adaptive generation for downstream application.

\vspace{1mm}
\noindent\textbf{Acknowledgements.} This material is based upon work supported by the Defense Advanced Research Projects Agency (DARPA) under Contract No. FA8750-19-C-1001. Any opinions, findings and conclusions or recommendations expressed in this material are those of the author(s) and do not necessarily reflect the views of the Defense Advanced Research Projects Agency (DARPA). This work was also supported by Army Research Office Grant W911NF2110246, National Science Foundation grants CCF-2007350 and CCF-1955981, and the Hariri institute at Boston University. We would also like to thank the developers of TDW: Seth Alter, Abhishek Bhandwaldar and Jeremy Schwartz, for their assistance with the platform and its use.

%% file: latex/appendix.tex
\appendix

\section*{Appendices}

\section{Task2Vec}
We used Task2Vec~\cite{achille2019task2vec} representations for downstream tasks for our Task2Sim model. Task2Vec of a task consists of diagonal elements of the Fisher information matrix (FIM) of the outputs with respect to the network parameters over the distribution of downstream task examples (Refer to Section 2 of \cite{achille2019task2vec} for more details). For this purpose, following \cite{achille2019task2vec}, we used a single Imagenet pre-trained probe network with only the classifier layer trained on specific tasks (using the training set of examples for that task). In our experiments, a Resnet-18 probe network was used, resulting in a 9600-dimensional Task2Vec task representation.

\vspace{1mm}
\noindent \textbf{How much downstream data do we need access to?} In the case of models pre-trained using an approach that is not task-adaptive, there is no need to access any downstream data while pre-training. Given that task-adaptive approaches need a downstream task representation we used Task2Vec. Here, following~\cite{achille2019task2vec} we used all labeled examples from the training set of the downstream task to represent its distribution (in computing the FIM). However, we show that the FIM can be estimated by using fewer examples from the downstream task and the resulting Task2Vec vectors can be used to train Task2Sim with no degradation in performance (see \cref{tab:diff_task2vec}). This property also makes Task2Sim more practical since a user need not wait to collect labels for all data pertaining to their downstream application in order to generate pre-training data using Task2Sim.

\begin{table}[]
    \centering
    \begin{tabular}{c|cc}
    \multicolumn{1}{c|}{\multirow{2}[1]{*}{\thead{\textbf{Fraction of data} \\ \textbf{used for Task2Vec}}}} & \multicolumn{2}{c}{\textbf{Avg Downstream Acc.}} \\
          & Seen Tasks & Unseen Tasks \\
    \midrule
    100\% & 30.46 & 53.06 \\
    50\%  & 30.69 & 52.70 \\
    20\%  & 30.72 & 53.11 \\
    10\%  & 31.18 & 53.57 \\
    \end{tabular}%
    \caption{Average downstream performance (evaluated with 5NN classifier and using 40k images from 100 classes for pre-training) over seen and unseen tasks using different fractions of downstream training data (randomly subsampled) used to compute Task2Vec task representations for Task2Sim model. Task2Sim performance does not degrade when fewer downstream examples are used for computing Task2Vec.}
    \label{tab:diff_task2vec}
\end{table}

\section{Similarity between Learned Features}
\input{figures/supp/cka}

We used centered kernel alignment (CKA)~\cite{kornblith2019similarity} to find the similarity between features learned by the Resnet-50 backbone pre-trained on different image sets containing 100k images from 237 classes. \cref{fig:cka}, shows these similarities computed using the output features at different stages of the backbone (Stages 1-4 are intermediate outputs after different convolutional blocks in the resnet).

A few interesting phenomena surface: Task2Sim features (\ie features produced by a model pre-trained on Task2Sim generated dataset) are more similar to Imagenet features, than Domain Randomization. Thus Task2Sim in some manner, mimics features learned on real images better. 
We can also see that features early on in the network are largely similar across all kinds of pre-training and they only start differentiating at later stages, suggesting high similarity in lower level features (\eg edges, curves, textures, \etc) across different pre-training datasets. Also, as might be expected, features post downstream finetuning are more similar to each other than before, while still quite far away from being identical.

\section{Additional Results}

\subsection{Effect of Different Backbones} \label{subsec:diff_backbones}

\input{figures/supp/diff_backbones_seen}
\input{figures/supp/diff_backbones_unseen}

In \cref{fig:diff_backbones_seen,,fig:diff_backbones_unseen}, we show the average downstream performance over the seen and unseen tasks respectively, using different Resnet backbones (of different sizes). For this study, we used the same pre-training procedure across all backbones. We see that results are largely consistent with different backbones and for all of them Task2Sim performance is competitive with Imagenet pre-training and is much better than Domain Randomization. We also see that typically methods improve average downstream performance with the use of a larger backbone in the classifier. Moving from Resnet-50 to Resnet-101, Task2Sim performance breaks this trend and is lower indicating that the larger backbone could overfit in this case. This might be expected since Task2Sim was trained to optimize the performance of a Resnet-50 backbone.

\subsection{Task2Sim Results---Linear Probing} 

\input{figures/supp/main_lineval}

\cref{fig:main_lineval} shows the downstream accuracy with linear probing for different seen and unseen datasets where pre-training dataset has 100k images from all 237 classes. These complement \cref{fig:main_all_seen,,fig:main_all_unseen}, where downstream evaluation used full network finetuning. 

\subsection{Varying Pre-training Data Size}

\input{figures/supp/diff_num_cls_lineval}

\input{figures/supp/diff_num_objs_lineval}

\input{figures/supp/diff_num_imgs_lineval}

\noindent \textbf{Linear Probing.} \cref{fig:diff_num_cls_lineval,,fig:diff_num_objs_lineval,,fig:diff_num_imgs_lineval} are counterparts (with downstream evaluation done with linear probing) of \cref{fig:diff_num_cls,,fig:diff_num_objs,,fig:diff_num_imgs} respectively. We see that primarily similar findings as the main paper hold and in \cref{fig:diff_num_cls_lineval}, different backbones improve at a similar rate with more classes (and images for pre-training). In \cref{fig:diff_num_objs_lineval}, we see that both methods of synthetic pre-training improve their features with more object models, with Domain Randomization improving at a slightly higher rate. 

In \cref{fig:diff_num_imgs_lineval} we see some differences: There is a more severe saturating behavior of downstream performance, which even decreases by a little after a certain point for the synthetic pre-training data. This is likely because the feature extractor overfits to the pre-training task and a linear classifier on these features cannot perform as well. Both from \cref{fig:diff_num_cls_lineval} and from the curve for Imagenet-1K in \cref{fig:diff_num_imgs_lineval} we see that this saturating/overfitting behavior is somewhat alleviated by more classes in pre-training data. Another observation of note in \cref{fig:diff_num_imgs_lineval} is that the feature extractor pre-trained on Domain Randomization starts overfitting \emph{before} it matches the performance of Task2Sim. 
With \cref{fig:diff_num_imgs}, we mentioned that with more images a non-adaptive approach like domain randomization could improve its performance faster and sometimes equal a task-adaptive approach like Task2Sim. \cref{fig:diff_num_imgs_lineval} shows that although a non-adaptive approach may improve faster, it may not always match performance of its adaptive counterpart.

\vspace{1mm}
\noindent \textbf{Unseen Tasks.} \cref{fig:diff_num_cls_unseen,,fig:diff_num_objs_unseen,,fig:diff_num_imgs_unseen} show effect of above variations averaged over unseen tasks. We can see that similar trends hold in this case, as in case of seen datasets.

\subsection{Comparison with Large scale Pre-training (CLIP)} CLIP~\cite{radford2021learning} pre-trains on ~400M image-text pairs. Such large datasets when curated from the web, are bound to have privacy and other ethical concerns, as discussed in the paper. CLIP pre-training is also much more expensive than its counterparts using our synthetic data. We conducted an experiment finetuning a Resnet-50 model using pre-trained weights from CLIP on our tasks, while noting that this CLIP pre-trained Resnet-50 is different from the standard model used by us and uses more parameters (38M in CLIP Resnet50 vs 25M in standard Resnet50). The result was 77.33\% avg. accuracy on seen tasks and 91.56\% avg. accuracy on unseen tasks, which is comparable to the best Task2Sim performance (79.10\% over seen tasks and 91.50\% over unseen tasks).

\section{Synthetic Image Generation} \label{sec:syn-data}

\input{figures/supp/sim_variations}

We used Three-D-World (TDW)~\cite{gan2020threedworld} for synthetic image generation. It is a platform built using the Unity3D engine, and besides a python interface, provides asset bundles which include 3D object models, interactive 3D scenes, and HDRI skyboxes (360$^\circ$ images of real scenes accompanied with lighting information). TDW is available under a BSD 2-Clause "Simplified" License.

For our implementation, we used all 2322 object models from 237 different classes available in TDW. We use a generator that imports one object into a simple scene with an HDRI-skybox background. It then, changes different properties of the scene/object/camera based on 8 simulation parameters as mentioned in \cref{subsec:expts_details}. Whenever different variations corresponding to a simulation parameter are to be included, values are chosen uniformly at random within an appropriate range (via a careful choice of the extremes). \cref{fig:sim_variations} has 8 rows corresponding to each of the simulation parameters used for Task2Sim. Each row shows using 5 images, the variations corresponding to its specific simulation parameter. 
Generating 1M images using our generator with all 2322 objects, takes around 12 hours on an Nvidia Tesla-V100 GPU. Given the number of objects we used in our implementation, a bottleneck in image generation is the speed of loading object meshes into Unity3D. Hence, we used a subset of 780 objects from 100 classes with relatively simpler meshes, for generating the data used for training Task2Sim. The 8 parameters we used result in a total of $2^8 = 256$ different possibilities and so we pre-generated these 256 sets of 40k images each for faster and smoother training of the Task2Sim model. Each of these 256 sets took $\sim$30 mins to generate on a Tesla-V100 GPU.

\section{Training and Evaluation} \label{sec:train_details}
We based our implementation of different classifiers for pre-training and downstream evaluation on pytorch-image-models~\cite{rw2019timm}. For all experiments except those in \cref{subsec:diff_backbones}, we used a Resnet-50 backbone for our classifier. For all datasets while pre-training, we used the following parameters: we trained for 100 epochs using an AdamW optimizer, using a learning rate 0.001 and a batch size of 1024. The learning rate used a linear warmup for 20 epochs and a cosine annealing schedule following warmup. We use regularization methods like label-smoothing, cutmix~\cite{yun2019cutmix} and mixup~\cite{zhang2017mixup} following a training strategy from \cite{rw2019timm}. We used image augmentation in the form of RandAugment~\cite{cubuk2020randaugment} while pre-training.

For downstream evaluation, we followed a procedure similar to \cite{islam2021broad}. For both evaluations using linear probing and full-network finetuning, we used 50 epochs of training using an SGD optimizer with learning rate decayed by a tenth at 25 and 37 epochs. No additional regularizers or data augmentation approaches were used. For each downstream task, we did a coarse hyperparameter grid-search over learning rate $\in \{10^{-5}, 10^{-4}, 10^{-3}, 10^{-2}\}$, optimizer weight decay $\in \{0, 10^{-5}\}$ and training batch size $\in \{32, 128\}$. We found by comparing backbones pre-trained on Imagenet and a large synthetic set generated with Domain Randomization, that with the above grid, for each specific downstream task and evaluation method, a particular set of hyperparameters worked best irrespective of the pre-training data. This was found using a separate validation split created from the downstream training set with 30\% of the examples. Given this finding, we fixed these hyperparameters for a given downstream task and evaluation method for all remaining experiments.

\section{Details of Downstream Tasks} \label{sec:downstream_details}
\cref{tab:dataset_details} shows the number of classes in each of the 20 downstream tasks we used. It also shows the number of images in the training and test splits for each.

\begin{table}[h]
    \centering
    \scalebox{0.8}{
    \begin{tabular}{l| l|r r r }
        \toprule
        Category & Dataset & Train Size & Test Size & Classes \\ 
        \midrule
        \multirow{3}{*}{Natural}
        & CropDisease~\cite{mohanty2016cropdisease} & 43456 & 10849 & 38 \\%&  \\
        & Flowers~\cite{nilsback2008automatedflowers102} & 1020 & 6149 & 102 \\%&  \\
        & DeepWeeds~\cite{DeepWeeds2019} & 12252 & 5257 & 9 \\%&  \\
        & CUB~\cite{WahCUB_200_2011} & 5994 & 5794 & 200 \\%&  \\
        \midrule
        \multirow{2}{*}{Satellite}
        & EuroSAT~\cite{helber2019eurosat} & 18900 & 8100 & 10 \\%&  \\
        & Resisc45~\cite{cheng2017remoteresisc45} & 22005 & 9495 & 45 \\%&  \\
        & AID~\cite{xia2017aid} & 6993 & 3007 & 30 \\%&  \\
        & CactusAerial~\cite{lopez2019columnarCactusAerial} & 17500 & 4000 & 2 \\%&  \\
        \midrule
        \multirow{2}{*}{Symbolic}
        & Omniglot~\cite{lake2015humanomniglot} & 9226 & 3954 & 1623 \\%&  \\
        & SVHN~\cite{netzer2011readingsvhn} & 73257 & 26032 & 10 \\%&  \\
        & USPS~\cite{hull1994databaseUSPS} & 7291 & 2007 & 10 \\%&  \\
        \midrule
        \multirow{2}{*}{Medical}
        & ISIC~\cite{codella2019skinisic} & 7007 & 3008 & 7  \\%&  \\
        & ChestX~\cite{wang2017chestx} & 18090 & 7758 & 7 \\%&  \\
        & ChestXPneumonia~\cite{kermany2018identifyingChestXP} & 5216 & 624 & 2 \\
        \midrule
        \multirow{2}{*}{Illustrative}
        & Kaokore~\cite{tian2020kaokore} & 6568 & 821 & 8 \\%&  \\
        & Sketch~\cite{wang2019learningsketch} & 35000 & 15889 & 1000 \\%&  \\
        & PACS-C~\cite{li2017deeperPACS} & 2107 & 237 & 7 \\%&  \\
        & PACS-S~\cite{li2017deeperPACS} & 3531 & 398 & 7 \\%&  \\
        \midrule
        \multirow{2}{*}{Texture}
        & DTD~\cite{cimpoi2014DTD} & 3760 & 1880 & 47 \\%&  \\
        & FMD~\cite{zhang2019poissonFMD} & 1400 & 600 & 10 \\%&  \\
        \bottomrule
    \end{tabular}
    }
    \caption{Number of classes in each downstream task and number of images in each training and test split.}
    \label{tab:dataset_details}
\end{table}

\input{figures/supp/diff_num_cls_unseen}

\input{figures/supp/diff_num_objs_unseen}

\input{figures/supp/diff_num_imgs_unseen}

\section{Limitations} \label{sec:limitations}
In this paper, we constrained our demonstration to a relatively low number of datasets and simulation parameters, limited by data generation, pre-training and evaluation speed. If these processes can be made more efficient, in future work, we can expect to use more simulation parameters (with possibly more discrete options or even real-valued ranges), and use more datasets for training \ours, allowing it to be more effective in deployment as a practical application.

While a large portion of contemporary representation learning research focuses on self-supervision to avoid using labels, we hope our demonstration with Task2Sim motivates further research in using simulated data from graphics engines for this purpose, with focus on adaptive generation for downstream application. 

\section{Societal Impact}
In the introduction, we discussed model pre-training using large real image datasets was what paved the way for a gamut of transfer learning research. Using real images is however riddled with curation costs and others concerns around privacy, copyright, ethical usage, etc. The fact that downstream performance on average correlates positively with the size of pre-training data, created a race for curating bigger datasets. Corporations with large resources are able to invest in such large-scale curation and create datasets for their exclusive use (\eg JFT-300M~\cite{chollet2017xception,hinton2015distilling}, or Instagram-3.5B~\cite{mahajan2018exploring}), which are unavailable to a range of research on downstream applications.

Using synthetic data for pre-training can drastically reduce these costs, because potentially infinite images can be rendered once 3D models and scenes are available, by varying various simulation parameters. In this paper, we demonstrated that the optimal use of such a simulation engine can be found in restricting certain variations, and that different restrictions benefit different downstream tasks. Our Task2Sim approach, can be used as the basis for a pre-training data generator, which as an end-user application can allow research on a wide range of downstream applications to have access to the benefits of pre-training on large-scale scale data. This does not create any direct impacts on average individuals, but could do so through the advancement in downstream applications. One particular case, as an example, could be the advancement in visual recognition systems in the medical domain, possibly making the diagnosis of illnesses faster and cheaper.

%% file: figures/supp/cka.tex
\begin{figure*}[t]
    \centering
    \begin{subfigure}[t]{\textwidth}
    \centering
    \includegraphics[width=\linewidth]{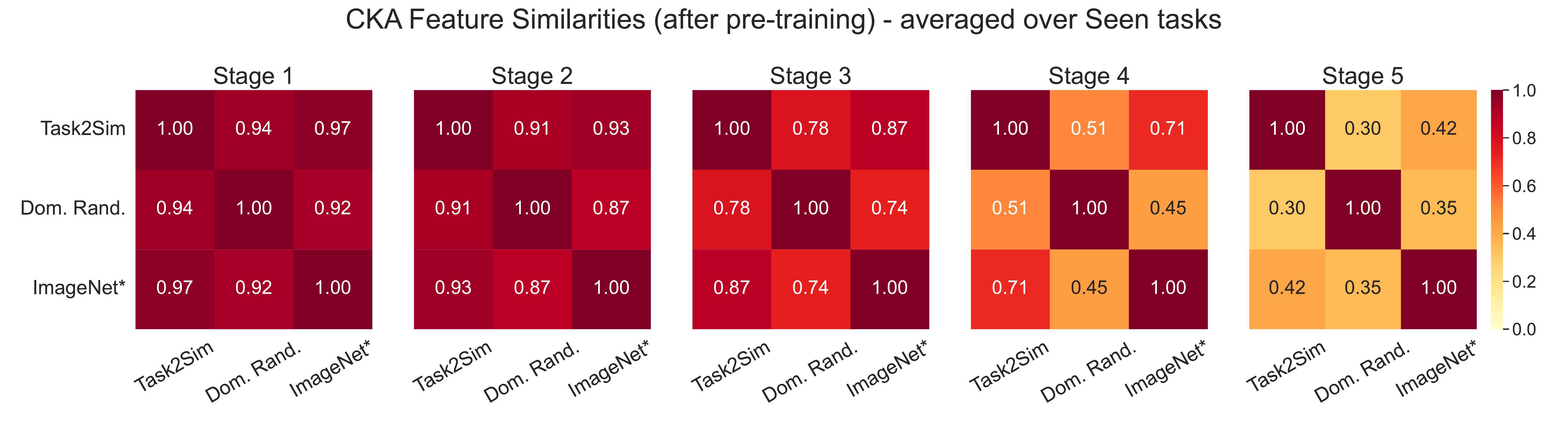}
    \vspace{-2mm}
    \end{subfigure}
    
    \begin{subfigure}[t]{\textwidth}
    \centering
    \includegraphics[width=\linewidth]{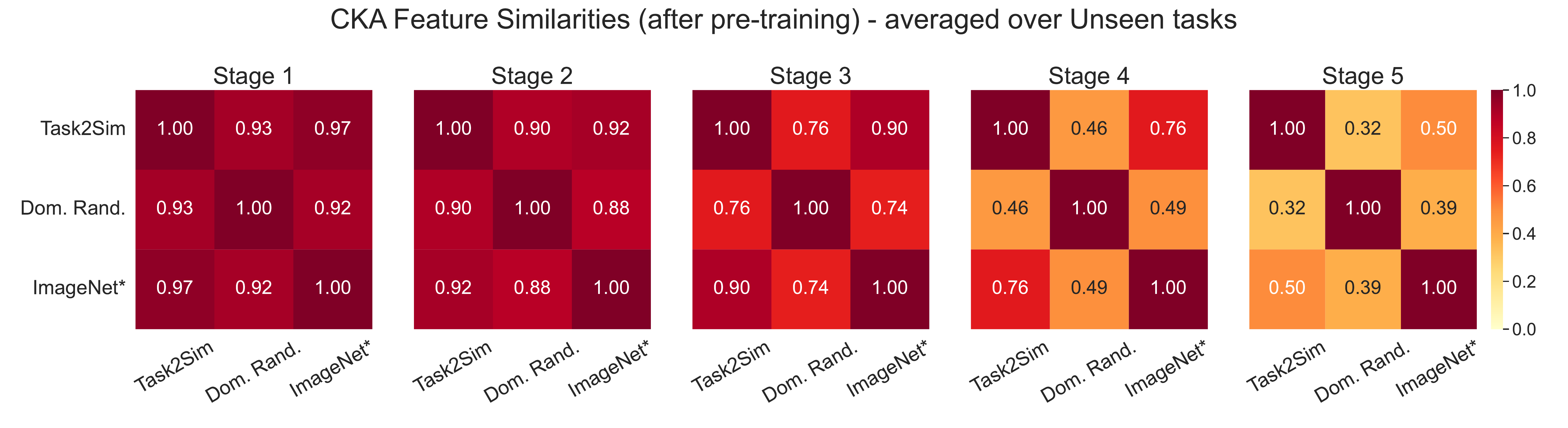}
    \vspace{-2mm}
    \end{subfigure}
    
    \begin{subfigure}[t]{\textwidth}
    \centering
    \includegraphics[width=\linewidth]{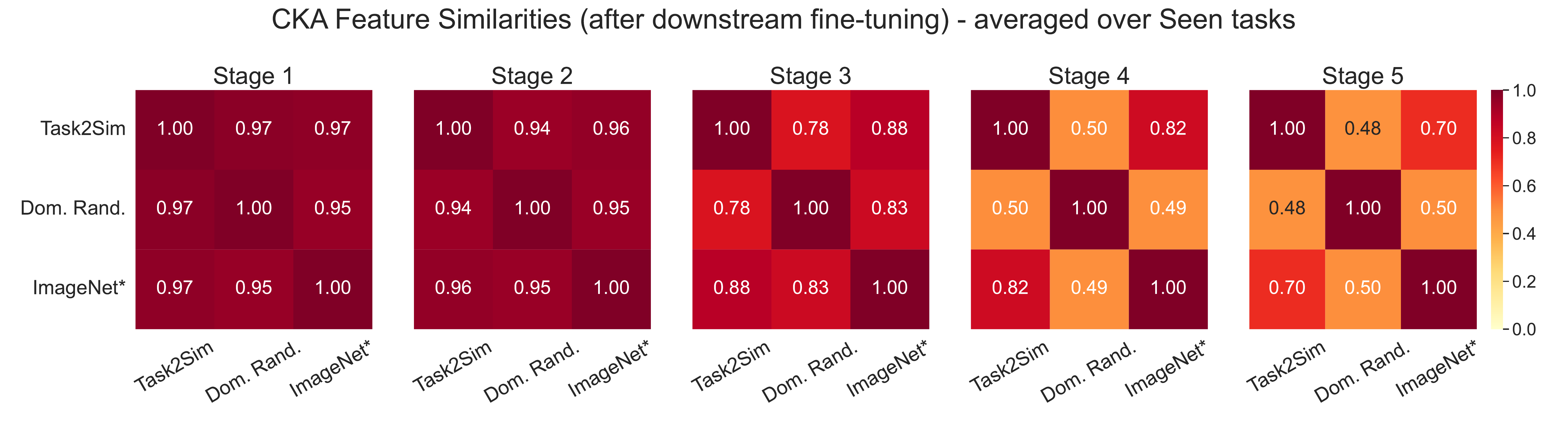}
    \vspace{-2mm}
    \end{subfigure}
    
    \begin{subfigure}[t]{\textwidth}
    \centering
    \includegraphics[width=\linewidth]{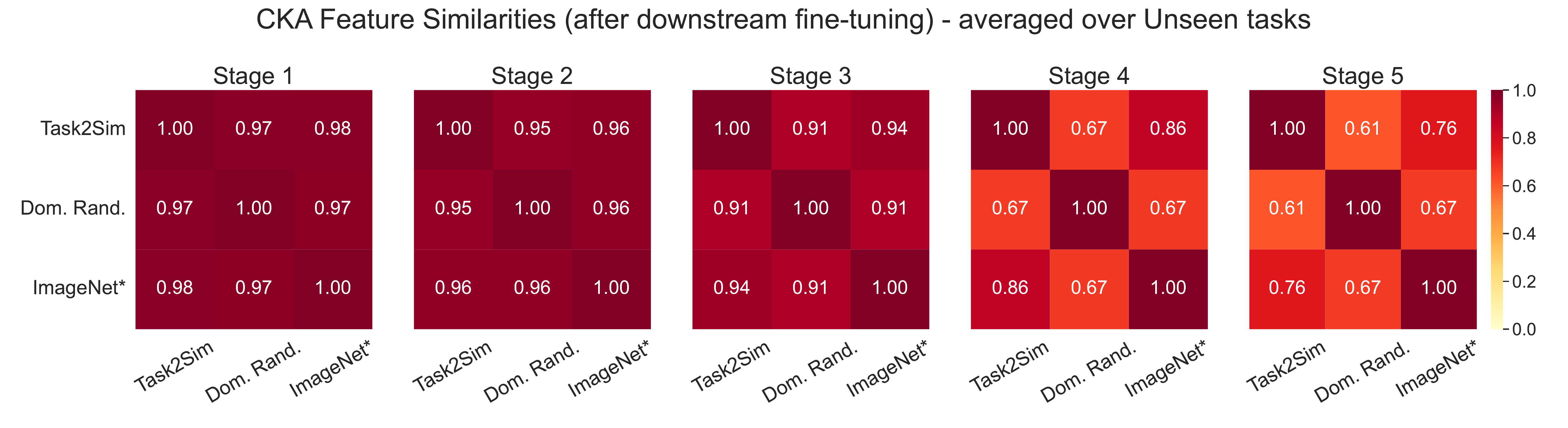}
    \vspace{-2mm}
    \end{subfigure}
    
    \caption{CKA similarities between features from backbones trained on different pre-training datasets (with 100k images from 237 classes). Similarities have been computed using features output at different stages of the Resnet-50 model. We notice that features at earlier stages across all methods of pre-training are quite similar and only later in the Resnet, do they start differentiating. We also observe that Task2Sim's features are more similar to Imagenet than those produced by pre-training with Domain Randomization.}
    \label{fig:cka}
    \vspace{-2mm}
\end{figure*}

%% file: figures/supp/diff_backbones_seen.tex
\begin{figure*}[t]
    \centering
    \begin{subfigure}[t]{0.48\textwidth}
    \centering
    \includegraphics[width=\linewidth]{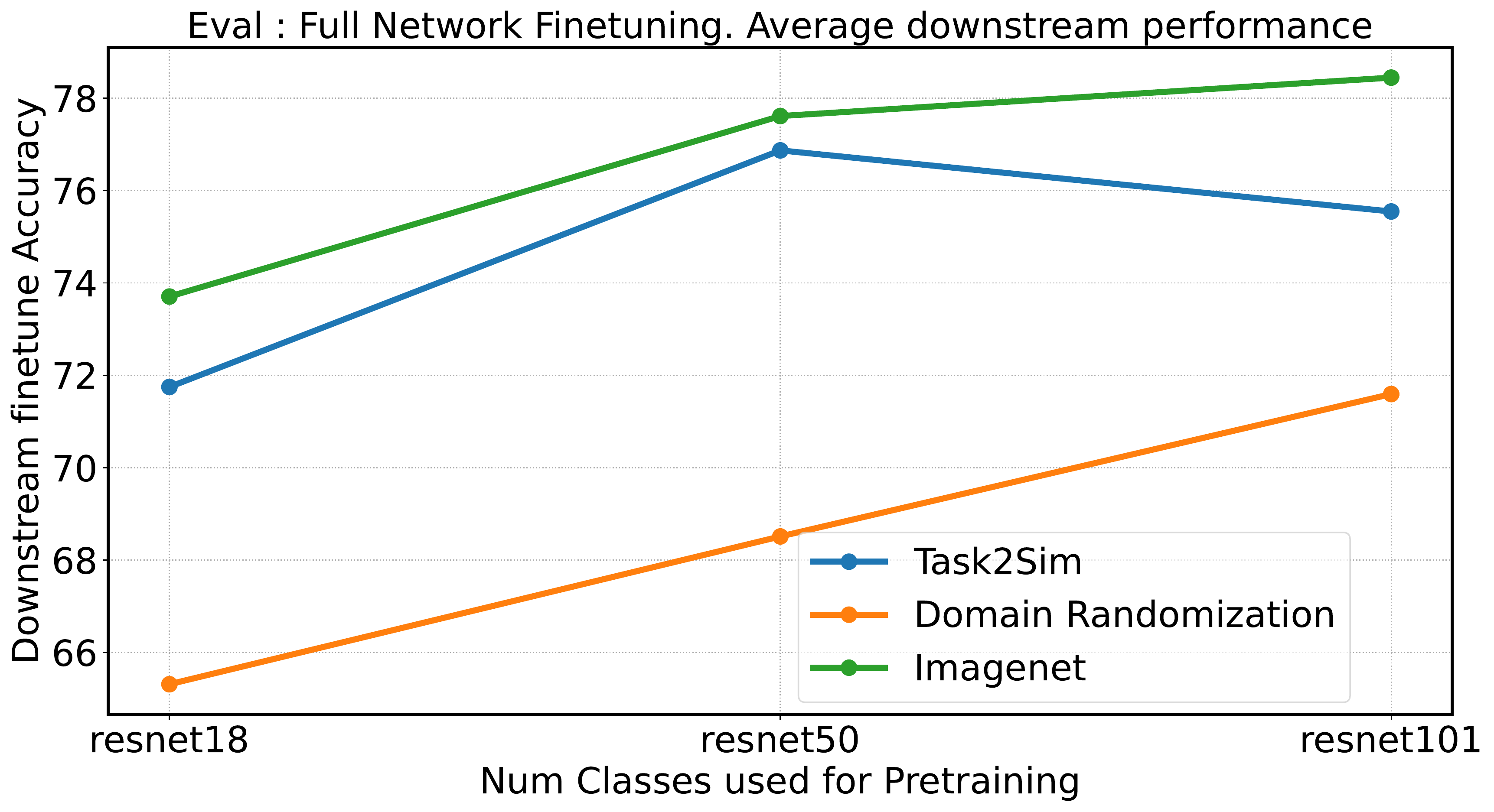}
    \vspace{-2mm}
    \end{subfigure}
    \begin{subfigure}[t]{0.48\textwidth}
    \centering
    \includegraphics[width=\linewidth]{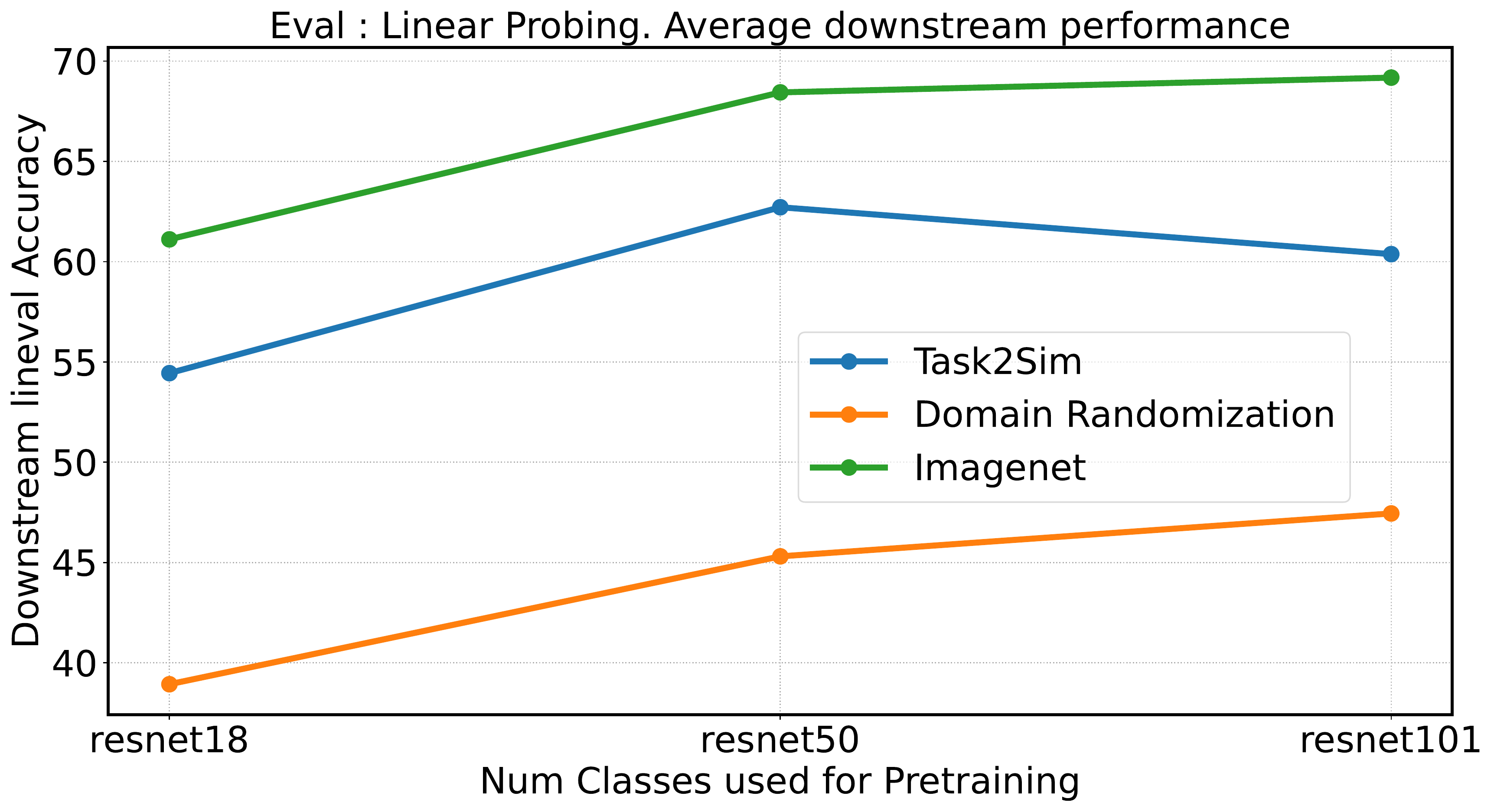}
    \end{subfigure}
    \vspace{-2mm}
    \caption{Effect of different backbones on average seen task performance (237 classes, 100k pre-training images). Best viewed in color.}
    \label{fig:diff_backbones_seen}
    \vspace{-2mm}
\end{figure*}

%% file: figures/supp/diff_backbones_unseen.tex
\begin{figure*}[t]
    \centering
    \begin{subfigure}[t]{0.48\textwidth}
    \centering
    \includegraphics[width=\linewidth]{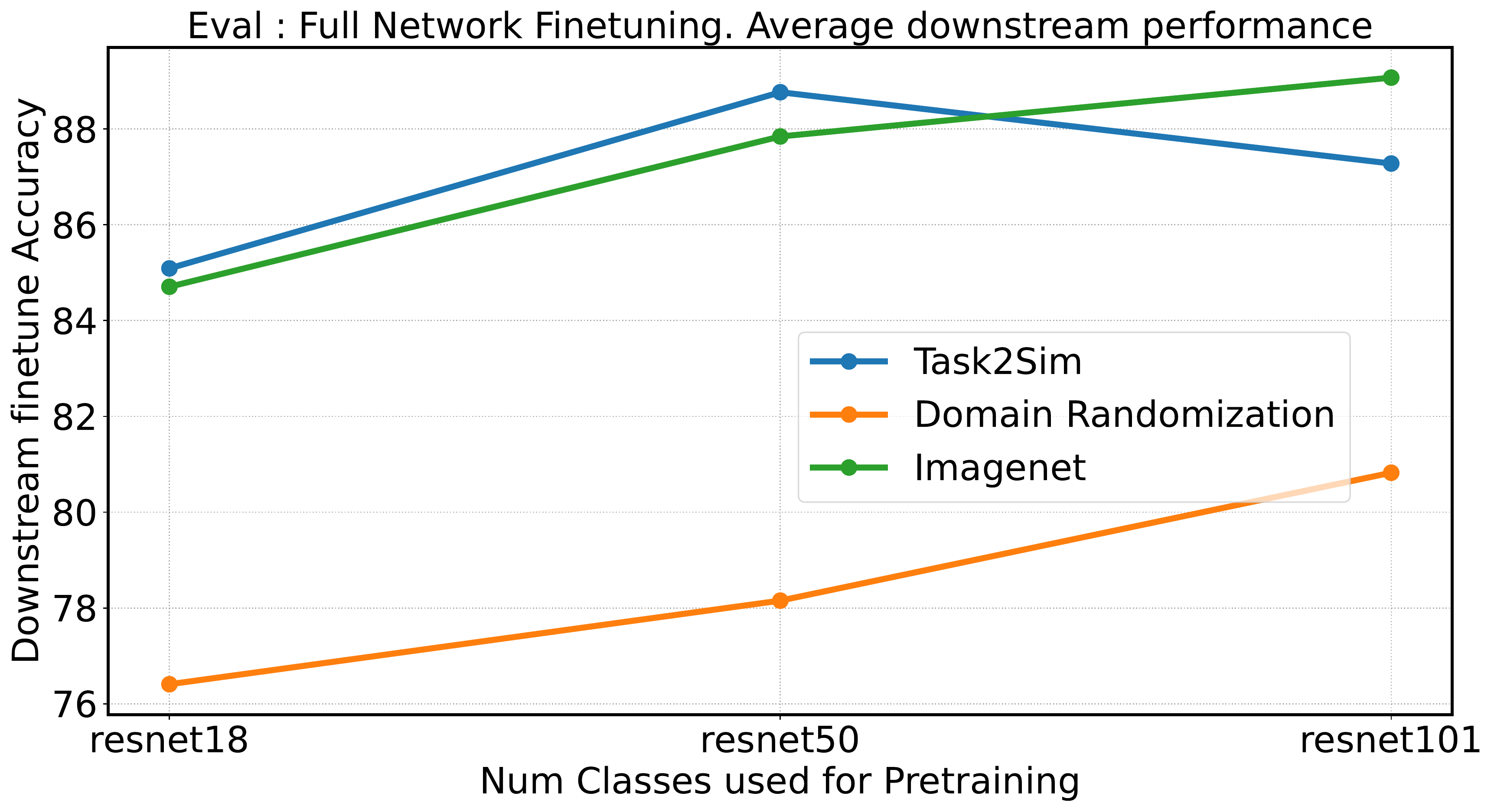}
    \vspace{-2mm}
    \end{subfigure}
    \begin{subfigure}[t]{0.48\textwidth}
    \centering
    \includegraphics[width=\linewidth]{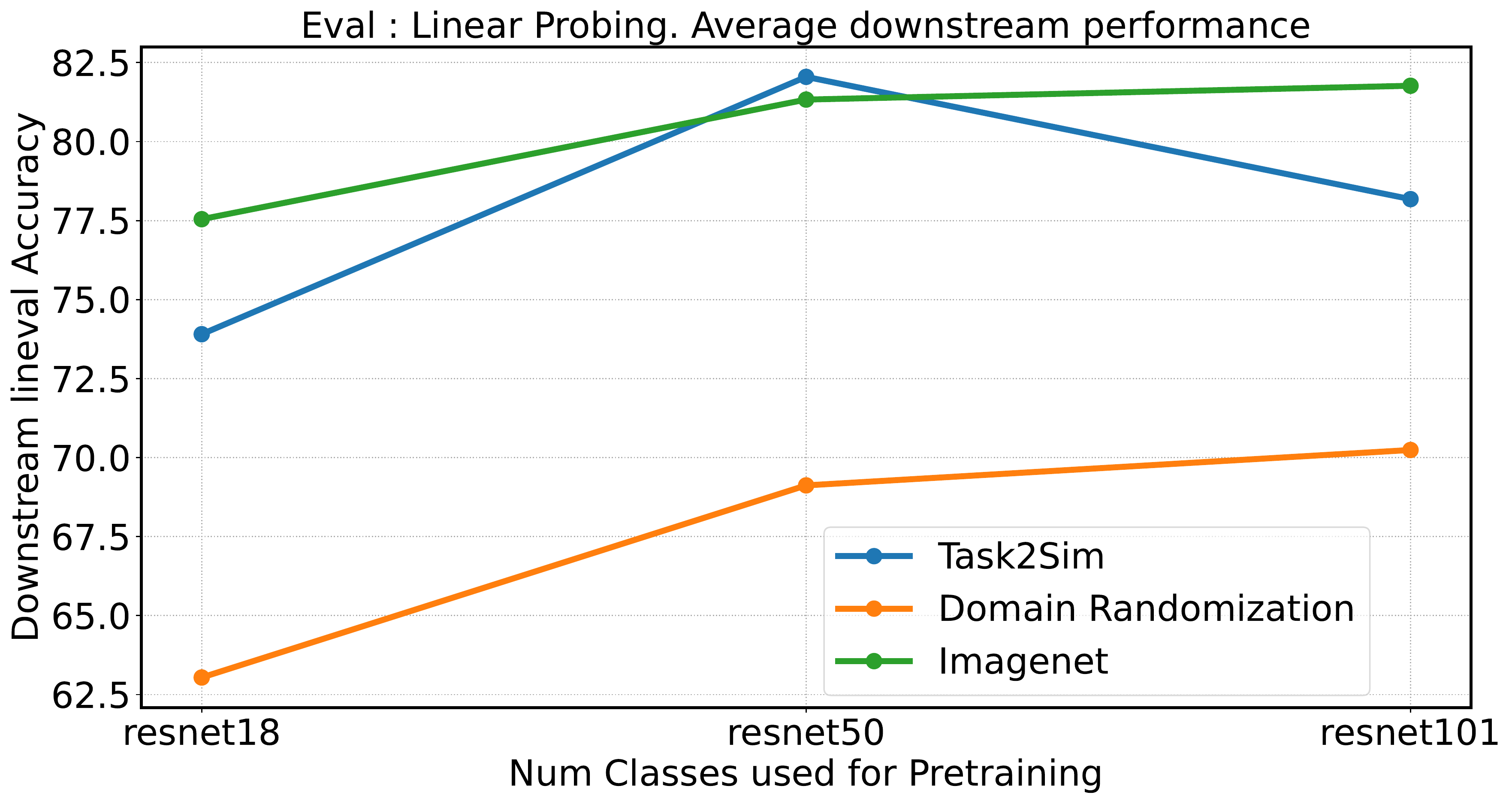}
    \end{subfigure}
    \vspace{-2mm}
    \caption{Effect of different backbones on average unseen task performance (237 classes, 100k pre-training images). Best viewed in color.}
    \label{fig:diff_backbones_unseen}
    \vspace{-2mm}
\end{figure*}

%% file: figures/supp/main_lineval.tex
\begin{figure*}[h]
    \centering
    \begin{subfigure}[t]{0.48\textwidth}
    \centering
    \includegraphics[width=\linewidth]{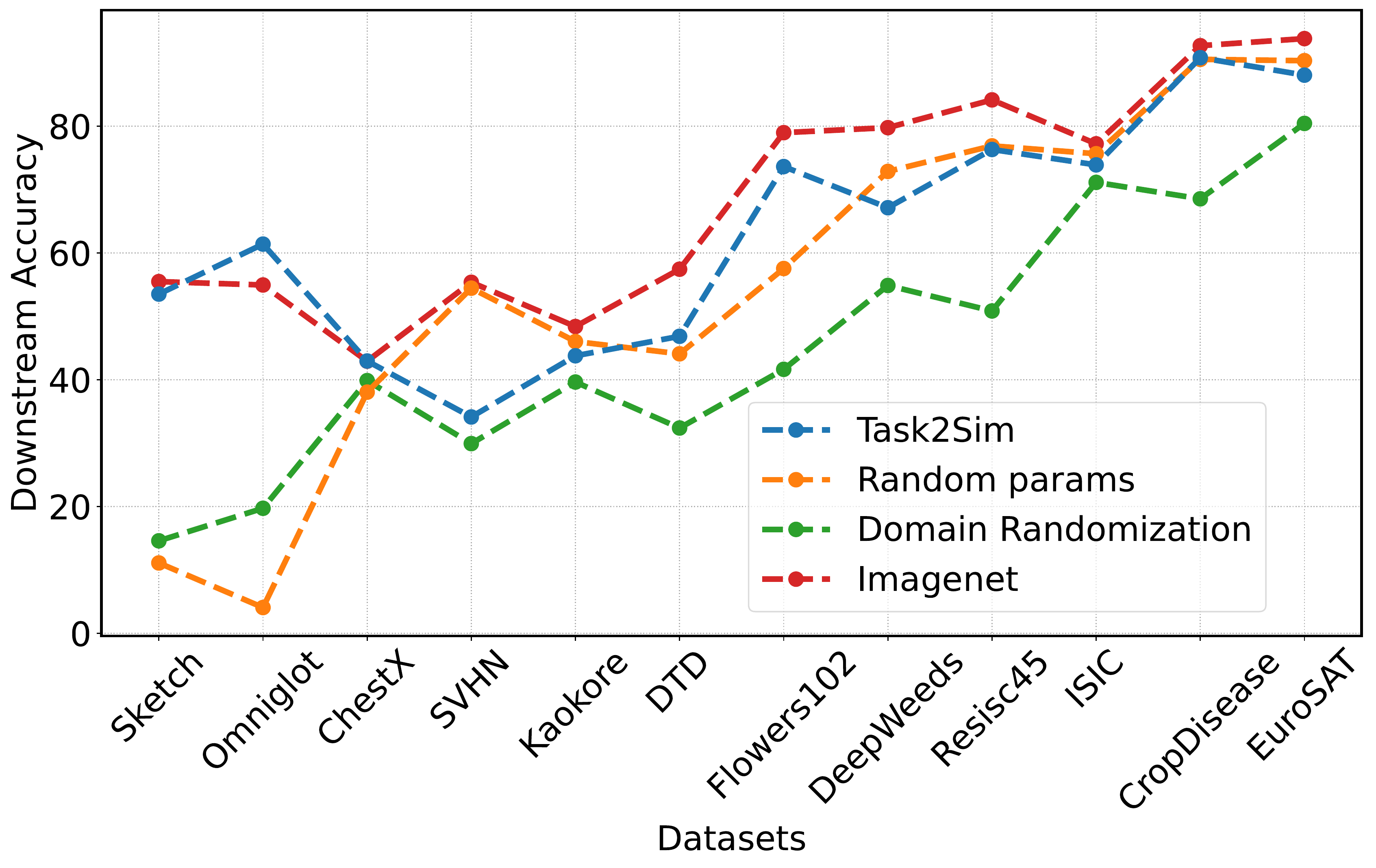}
    \vspace{-2mm}
    \end{subfigure}
    \begin{subfigure}[t]{0.48\textwidth}
    \centering
    \includegraphics[width=\linewidth]{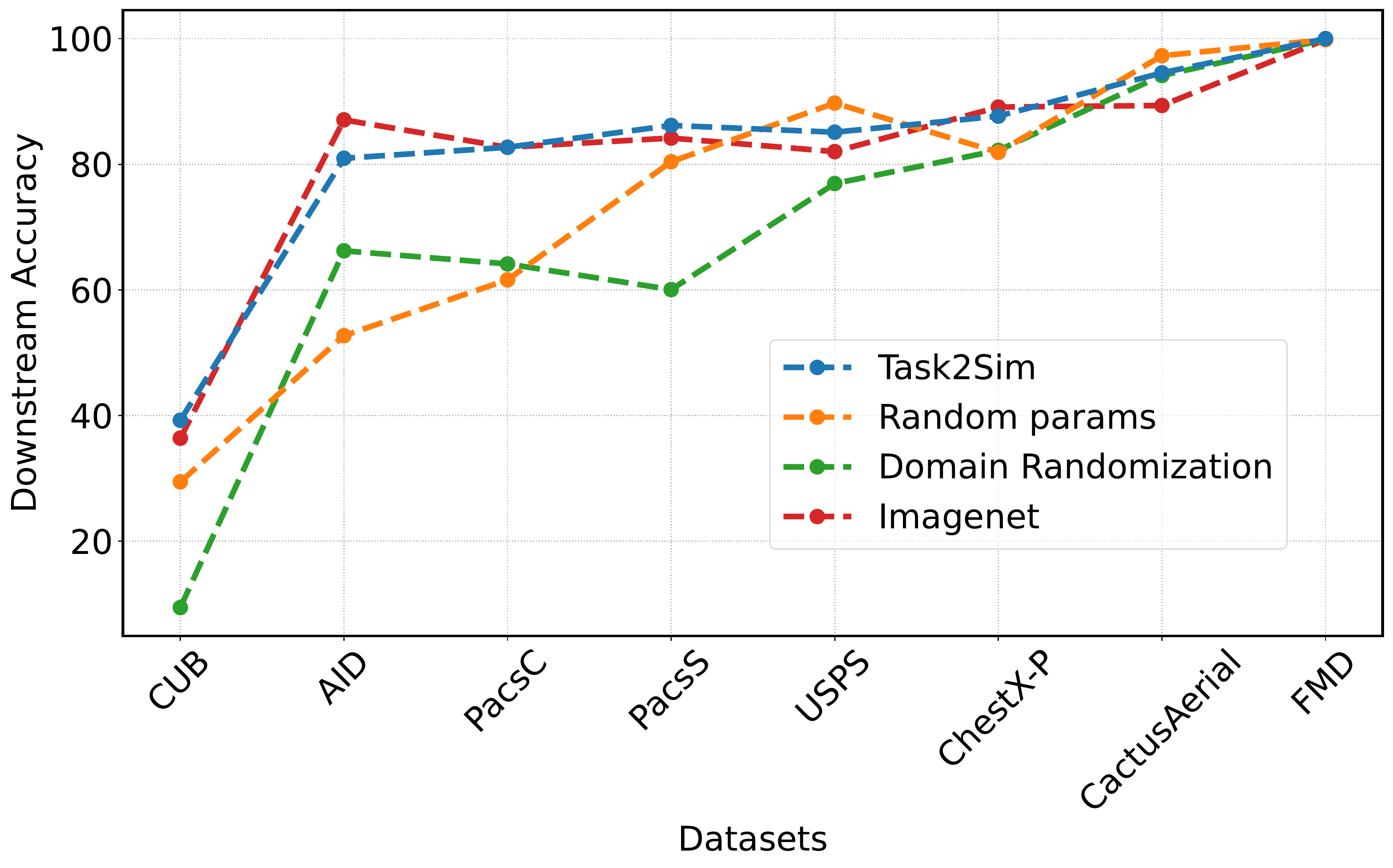}
    \end{subfigure}
    \vspace{-2mm}
    \caption{Performance of Task2Sim vs baselines on 12 seen tasks and 8 unseen tasks for 237 class / 100k image pre-training datasets evaluated with linear probing. Best viewed in color.}
    \label{fig:main_lineval}
    \vspace{-2mm}
\end{figure*}

%% file: figures/supp/diff_num_cls_lineval.tex
\begin{figure}[h]
    \centering
    \includegraphics[width=\linewidth,trim=0cm 0cm 0cm 0cm,clip]{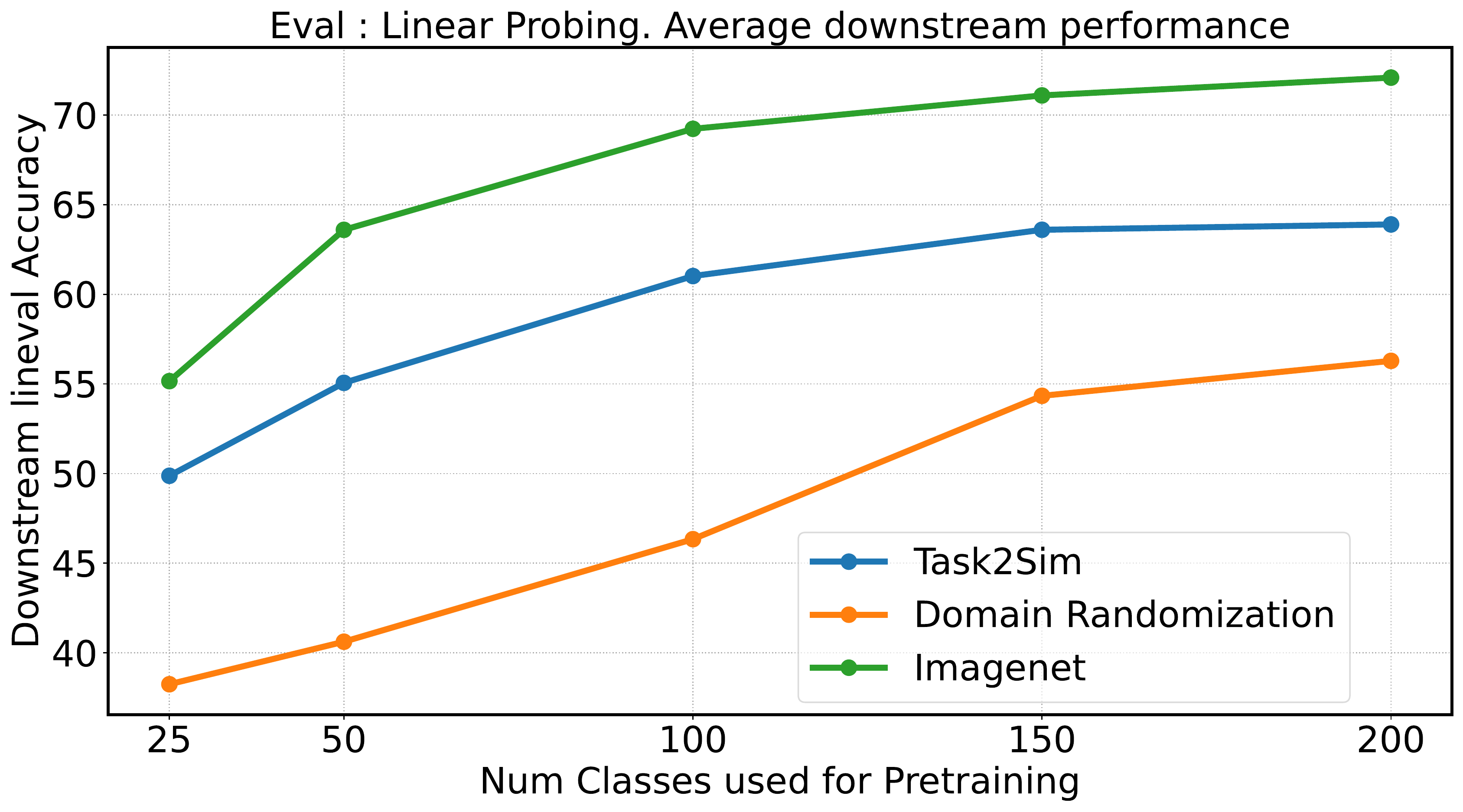}
    \caption{Avg performance with linear probing over 12 seen tasks at different number of classes for pre-training. All methods improve performance at similar rates with the addition of more classes.}
    \label{fig:diff_num_cls_lineval}
\end{figure}

%% file: figures/supp/diff_num_objs_lineval.tex
\begin{figure}[h]
    \centering
    \includegraphics[width=\linewidth,trim=0cm 0cm 0cm 0cm,clip]{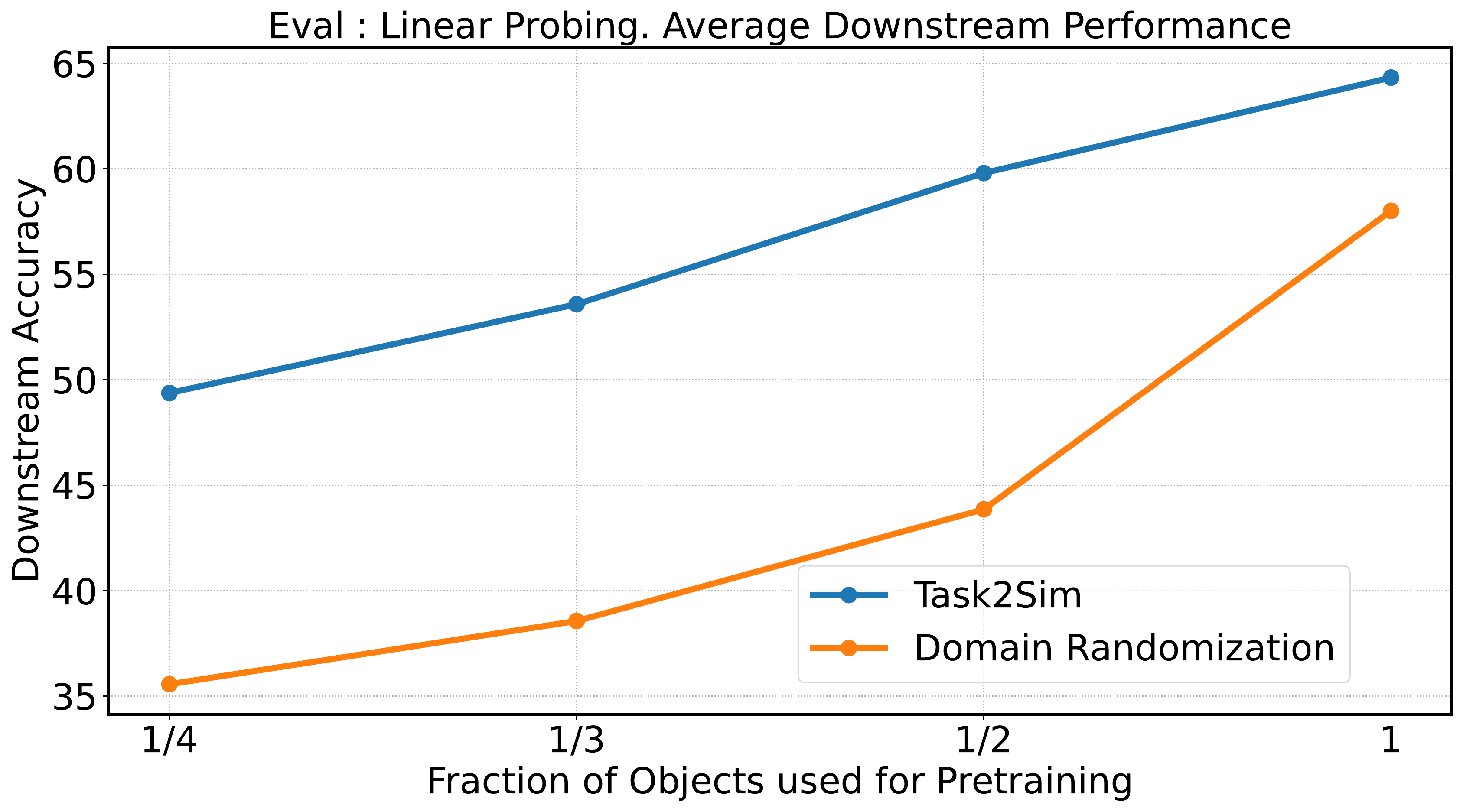}
    \caption{Avg performance with linear probing over 12 seen tasks at different number of object meshes used per category for generating synthetic pretraining data. Both methods of synthetic data generation improve performance with addition of more objects with Domain Randomization improving at a slightly higher rate.}
    \label{fig:diff_num_objs_lineval}
\end{figure}

%% file: figures/supp/diff_num_imgs_lineval.tex
\begin{figure}[h]
    \centering
    \includegraphics[width=\linewidth,trim=0cm 0cm 0cm 0cm,clip]{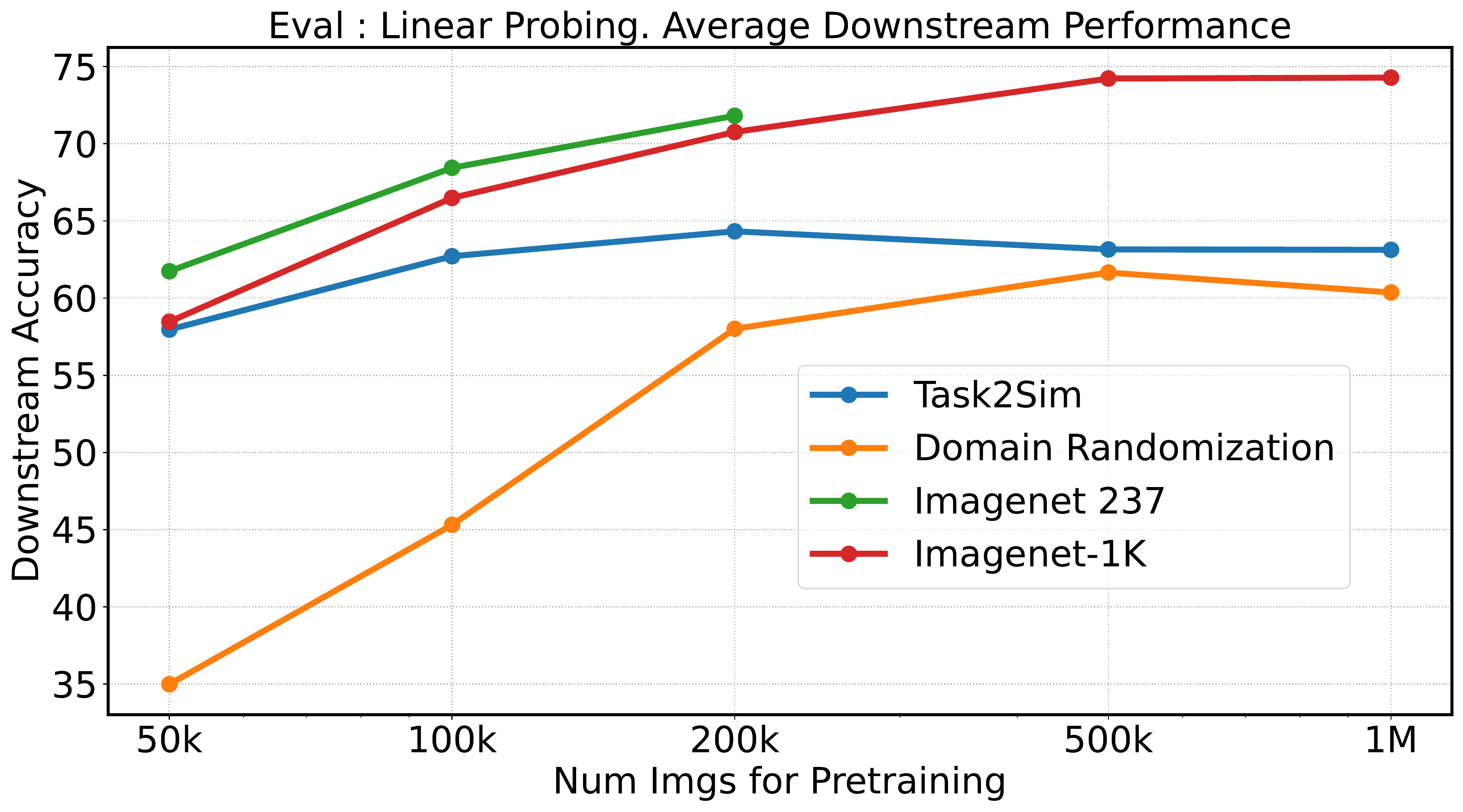}
    \caption{Task2Sim performance (avg over 12 seen tasks) vs other methods using linear probing for evaluation at different number of images for pretraining. Task2Sim is highly effective at fewer images. Increasing the number of images improves performance for all methods. Towards higher number of images in the case of linear probing we see methods not only reach a saturation but also exhibit some overfitting to pre-training data. Also, Domain Randomization stops improving in this case (evaluation with linear probing) before it can match Task2Sim performance.}
    \label{fig:diff_num_imgs_lineval}
\end{figure}

%% file: figures/supp/sim_variations.tex
\begin{figure*}[]
    \centering
    \includegraphics[width=0.95\linewidth,trim=0cm 0cm 0cm 0cm,clip]{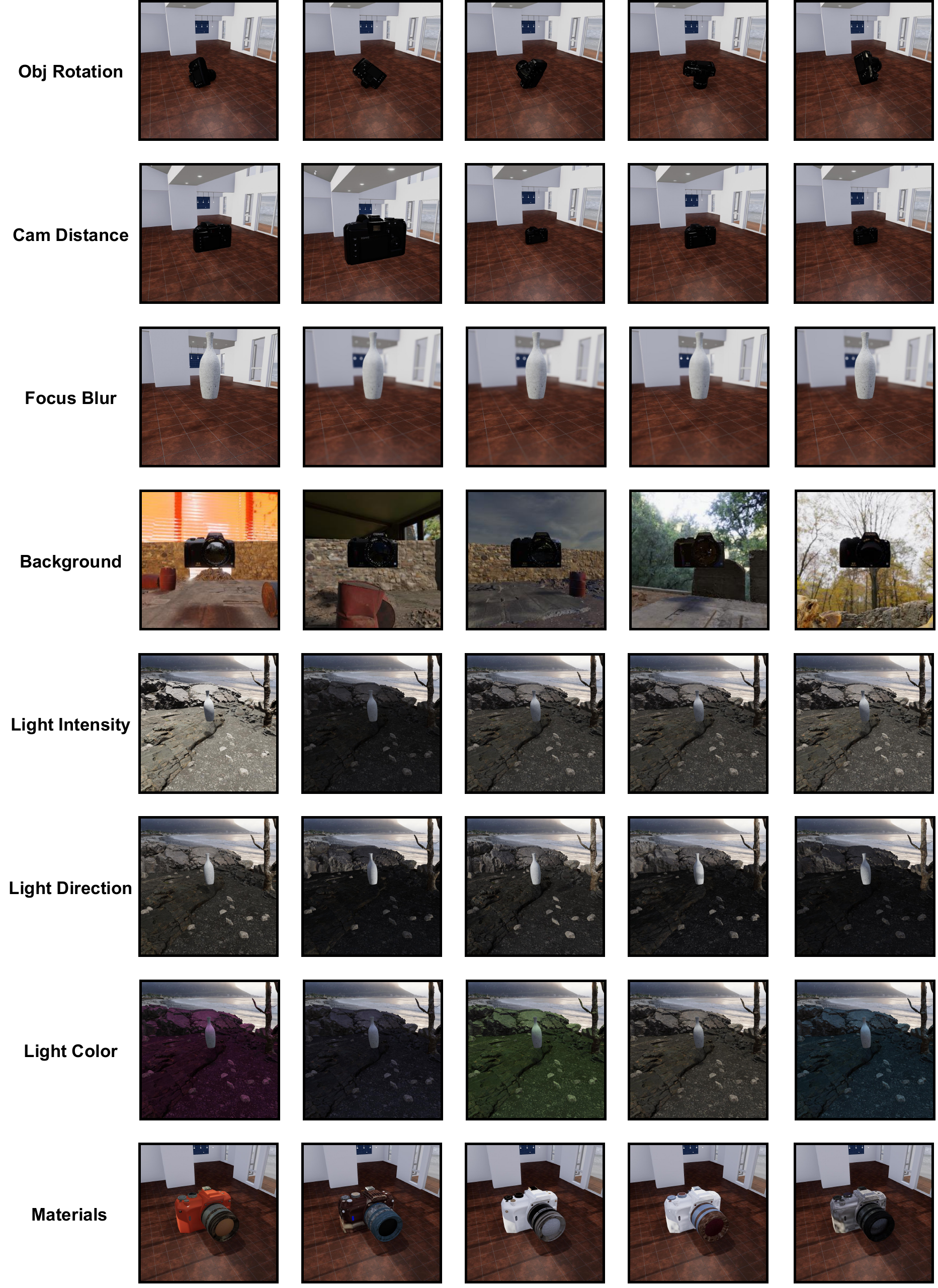}
    \vspace{-2mm}
    \caption{Examples of variations using different simulation parameters. Best viewed in color and under zoom.}
    \label{fig:sim_variations}
    \vspace{-2mm}
\end{figure*}

%% file: figures/supp/diff_num_cls_unseen.tex
\begin{figure*}[h]
    \centering
    \begin{subfigure}[t]{0.48\textwidth}
    \centering
    \includegraphics[width=\linewidth]{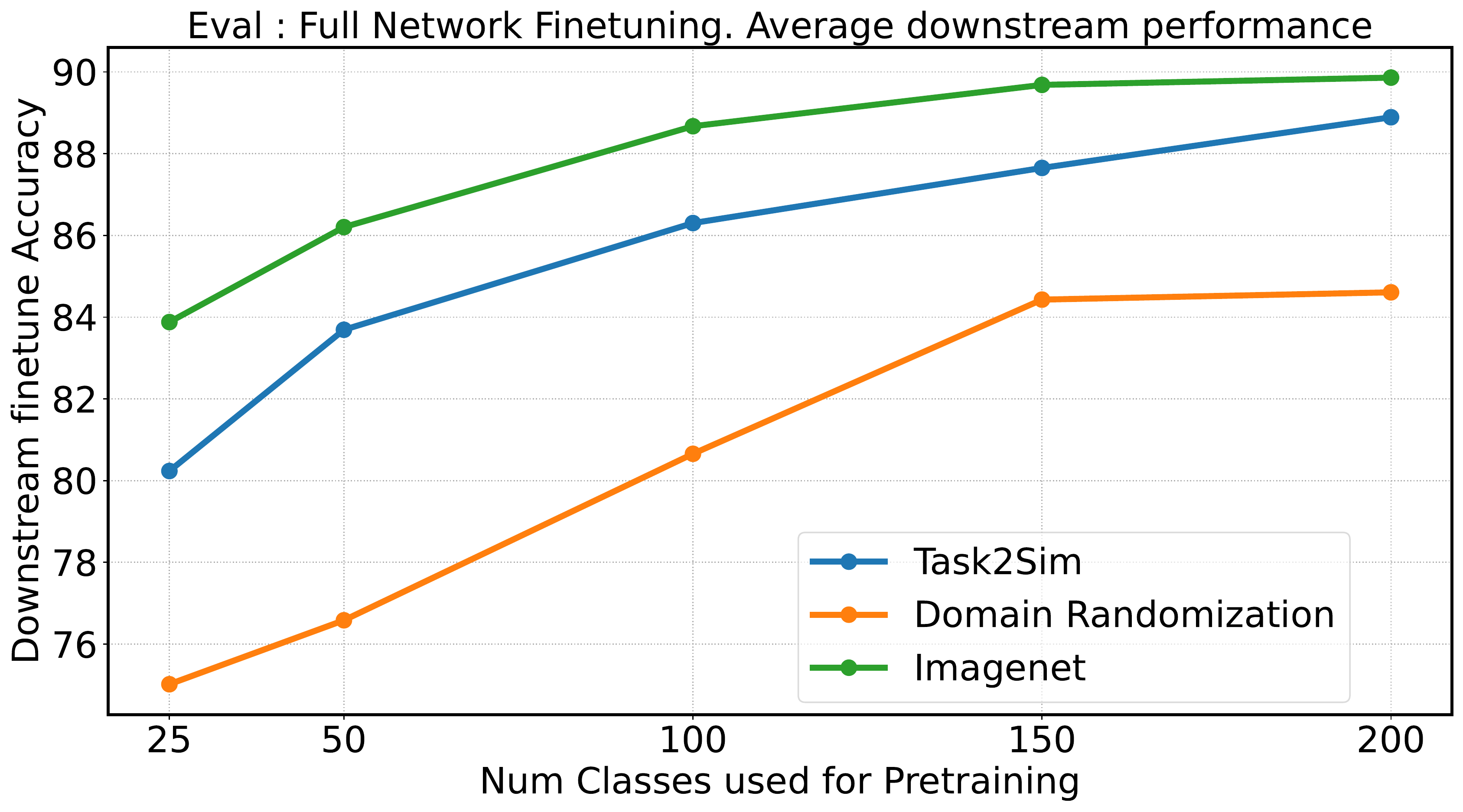}
    \vspace{-2mm}
    \end{subfigure}
    \begin{subfigure}[t]{0.48\textwidth}
    \centering
    \includegraphics[width=\linewidth]{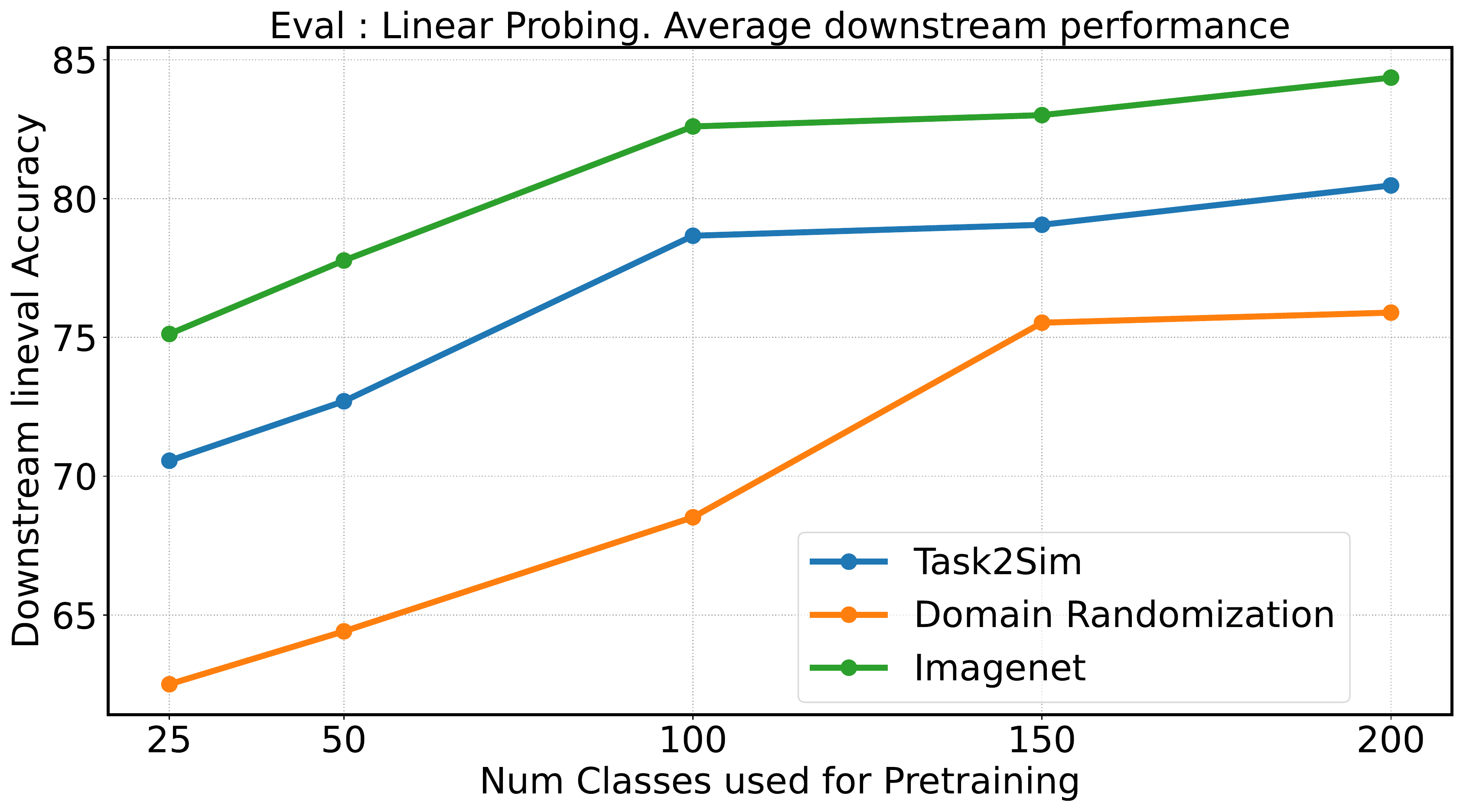}
    \end{subfigure}
    \vspace{-2mm}
    \caption{Downstream performance (avg over 8 unseen tasks) with different number of classes for pre-training. Best viewed in color.}
    \label{fig:diff_num_cls_unseen}
    \vspace{-2mm}
\end{figure*}

%% file: figures/supp/diff_num_objs_unseen.tex
\begin{figure*}[h]
    \centering
    \begin{subfigure}[t]{0.48\textwidth}
    \centering
    \includegraphics[width=\linewidth]{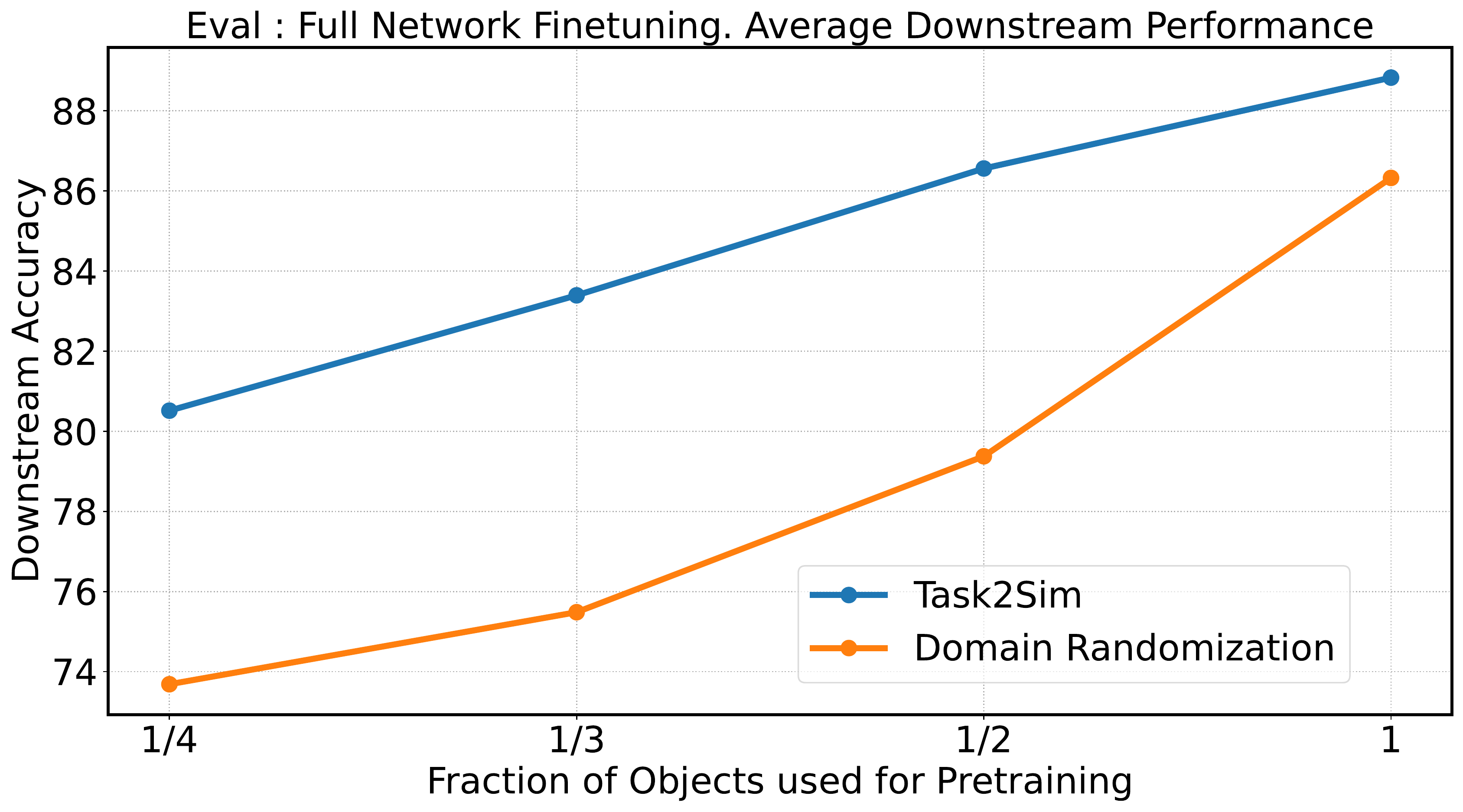}
    \vspace{-2mm}
    \end{subfigure}
    \begin{subfigure}[t]{0.48\textwidth}
    \centering
    \includegraphics[width=\linewidth]{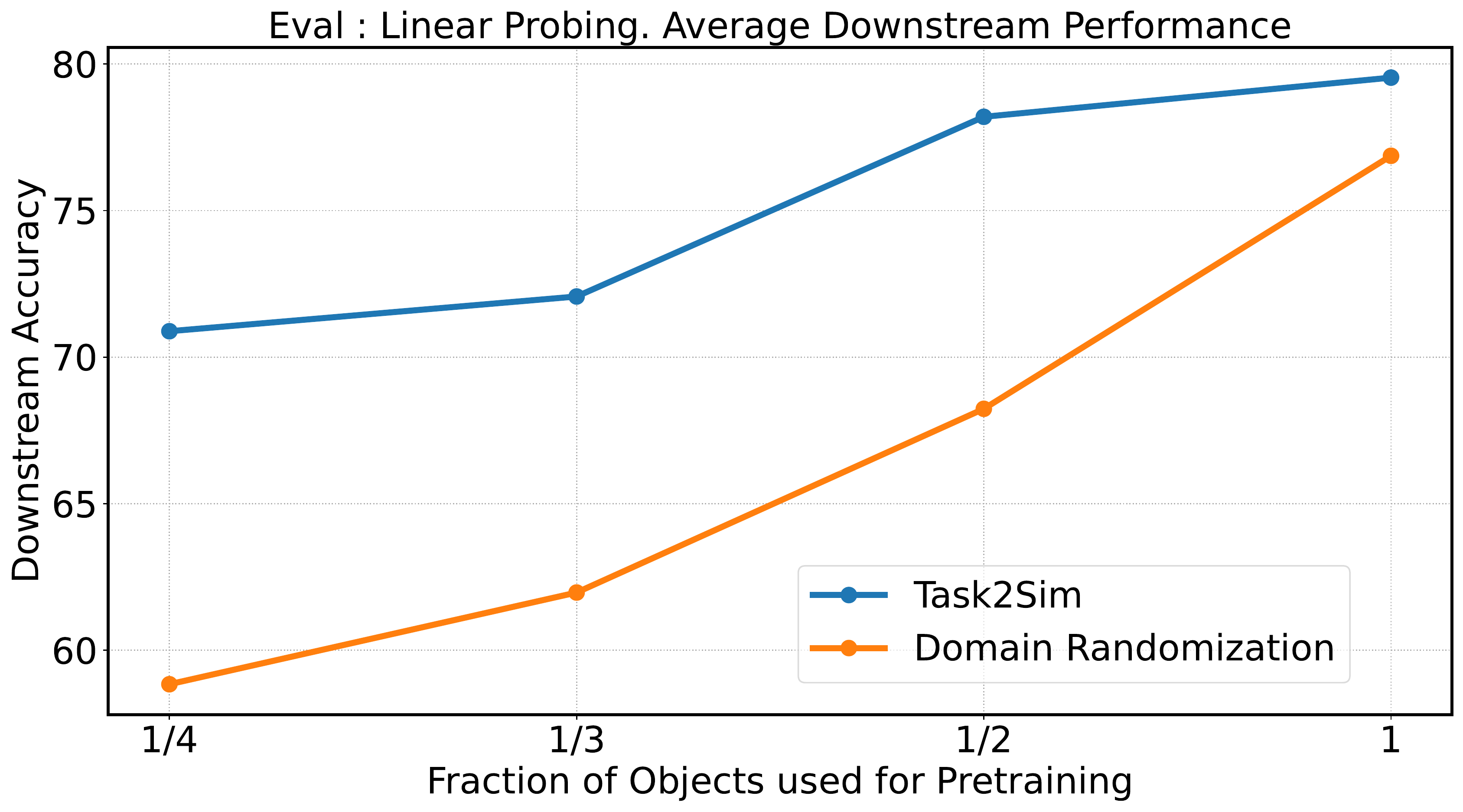}
    \end{subfigure}
    \vspace{-2mm}
    \caption{Downstream performance (avg over 8 unseen tasks) with different number of objects for pre-training. Best viewed in color.}
    \label{fig:diff_num_objs_unseen}
    \vspace{-2mm}
\end{figure*}

%% file: figures/supp/diff_num_imgs_unseen.tex
\begin{figure*}[h]
    \centering
    \begin{subfigure}[t]{0.48\textwidth}
    \centering
    \includegraphics[width=\linewidth]{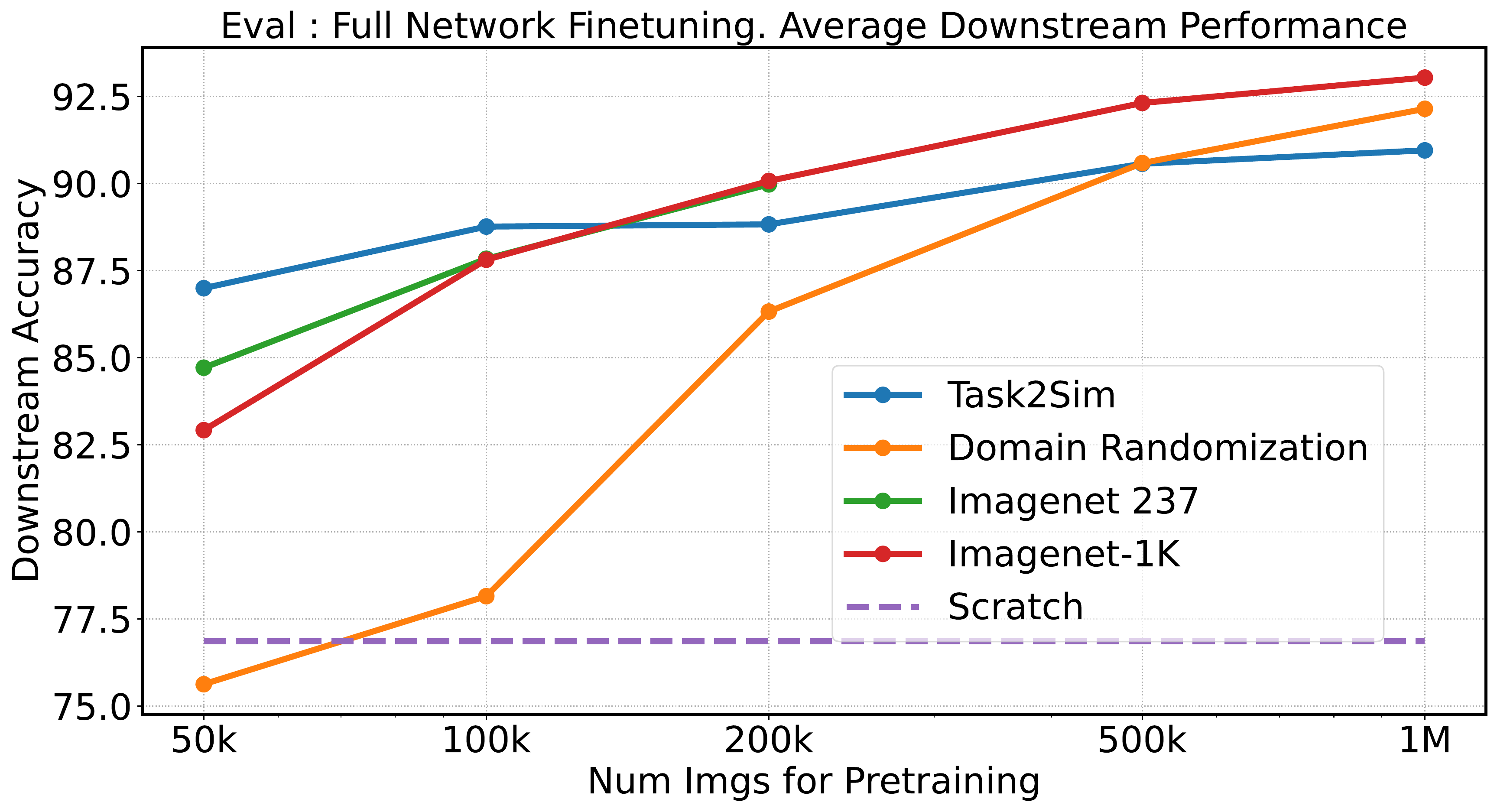}
    \vspace{-2mm}
    \end{subfigure}
    \begin{subfigure}[t]{0.48\textwidth}
    \centering
    \includegraphics[width=\linewidth]{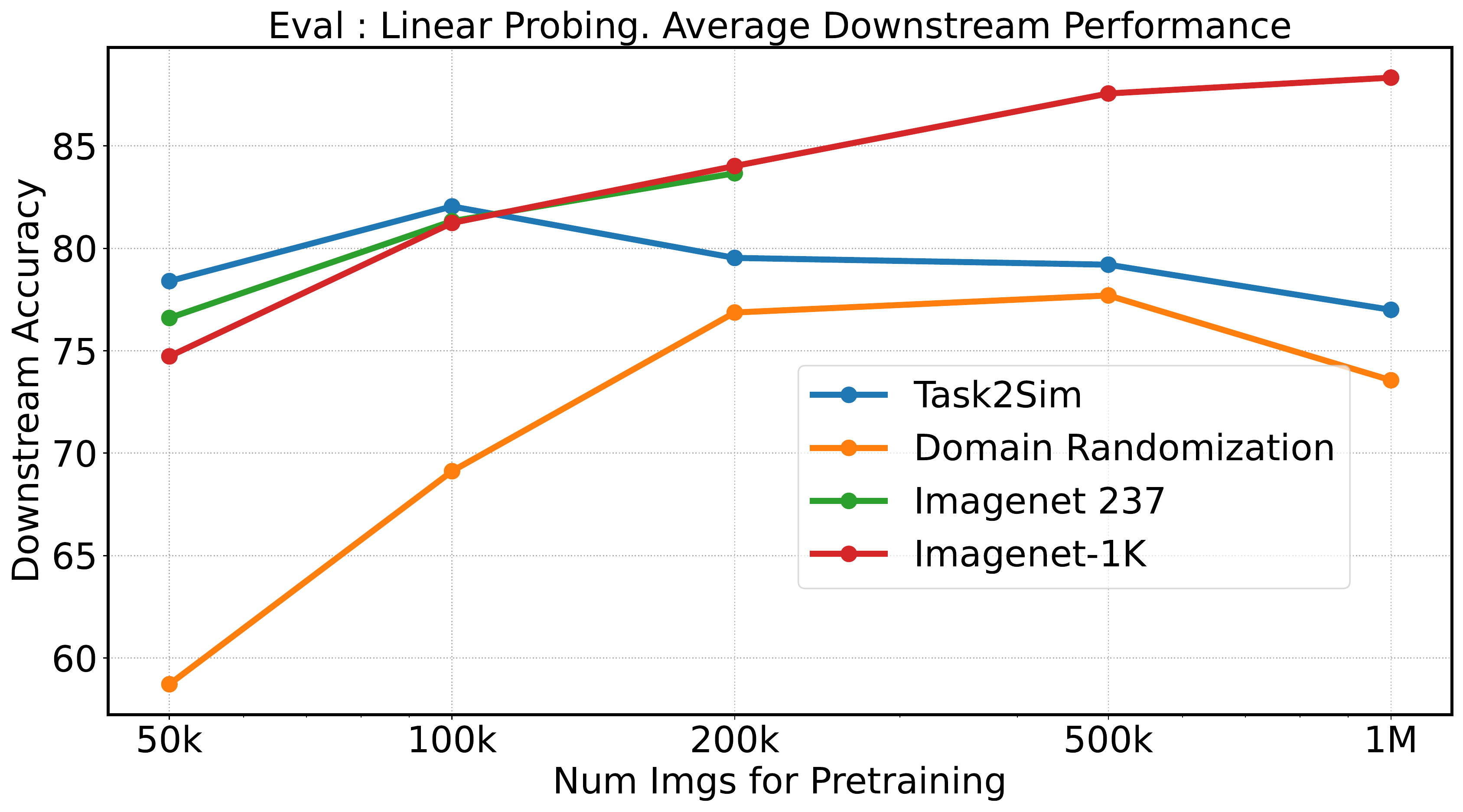}
    \end{subfigure}
    \vspace{-2mm}
    \caption{Downstream performance (avg over 8 unseen tasks) with different number of images for pre-training. Best viewed in color.}
    \label{fig:diff_num_imgs_unseen}
    \vspace{-2mm}
\end{figure*}

%% file: submission.bbl
\begin{thebibliography}{10}\itemsep=-1pt

\bibitem{achille2019task2vec}
Alessandro Achille, Michael Lam, Rahul Tewari, Avinash Ravichandran, Subhransu
  Maji, Charless~C Fowlkes, Stefano Soatto, and Pietro Perona.
\newblock Task2vec: Task embedding for meta-learning.
\newblock In {\em ICCV}, 2019.

\bibitem{anderson2021sim}
Peter Anderson, Ayush Shrivastava, Joanne Truong, Arjun Majumdar, Devi Parikh,
  Dhruv Batra, and Stefan Lee.
\newblock Sim-to-real transfer for vision-and-language navigation.
\newblock In {\em CoRL}, 2021.

\bibitem{baradad2021learning}
Manel Baradad, Jonas Wulff, Tongzhou Wang, Phillip Isola, and Antonio Torralba.
\newblock Learning to see by looking at noise.
\newblock {\em arXiv preprint arXiv:2106.05963}, 2021.

\bibitem{behl2020autosimulate}
Harkirat~Singh Behl, Atilim~G{\"u}ne{\c{s}} Baydin, Ran Gal, Philip~HS Torr,
  and Vibhav Vineet.
\newblock Autosimulate:(quickly) learning synthetic data generation.
\newblock In {\em ECCV}, 2020.

\bibitem{cheng2017remoteresisc45}
Gong Cheng, Junwei Han, and Xiaoqiang Lu.
\newblock Remote sensing image scene classification: Benchmark and state of the
  art.
\newblock {\em Proceedings of the IEEE}, 105(10):1865--1883, 2017.

\bibitem{chollet2017xception}
Fran{\c{c}}ois Chollet.
\newblock Xception: Deep learning with depthwise separable convolutions.
\newblock In {\em Proceedings of the IEEE conference on computer vision and
  pattern recognition}, pages 1251--1258, 2017.

\bibitem{cimpoi2014DTD}
Mircea Cimpoi, Subhransu Maji, Iasonas Kokkinos, Sammy Mohamed, and Andrea
  Vedaldi.
\newblock Describing textures in the wild.
\newblock In {\em Proceedings of the IEEE Conference on Computer Vision and
  Pattern Recognition}, pages 3606--3613, 2014.

\bibitem{codella2019skinisic}
Noel Codella, Veronica Rotemberg, Philipp Tschandl, M~Emre Celebi, Stephen
  Dusza, David Gutman, Brian Helba, Aadi Kalloo, Konstantinos Liopyris, Michael
  Marchetti, et~al.
\newblock Skin lesion analysis toward melanoma detection 2018: A challenge
  hosted by the international skin imaging collaboration (isic).
\newblock {\em arXiv preprint arXiv:1902.03368}, 2019.

\bibitem{csurka2017domain}
Gabriela Csurka.
\newblock Domain adaptation for visual applications: A comprehensive survey.
\newblock {\em arXiv preprint arXiv:1702.05374}, 2017.

\bibitem{cubuk2020randaugment}
Ekin~D Cubuk, Barret Zoph, Jonathon Shlens, and Quoc~V Le.
\newblock Randaugment: Practical automated data augmentation with a reduced
  search space.
\newblock In {\em Proceedings of the IEEE/CVF Conference on Computer Vision and
  Pattern Recognition Workshops}, pages 702--703, 2020.

\bibitem{deng2009imagenet}
Jia Deng, Wei Dong, Richard Socher, Li-Jia Li, Kai Li, and Li Fei-Fei.
\newblock Imagenet: A large-scale hierarchical image database.
\newblock In {\em 2009 IEEE conference on computer vision and pattern
  recognition}, pages 248--255. Ieee, 2009.

\bibitem{devaranjan2020meta}
Jeevan Devaranjan, Amlan Kar, and Sanja Fidler.
\newblock Meta-sim2: Unsupervised learning of scene structure for synthetic
  data generation.
\newblock In {\em ECCV}, 2020.

\bibitem{gan2020threedworld}
Chuang Gan, Jeremy Schwartz, Seth Alter, Martin Schrimpf, James Traer, Julian
  De~Freitas, Jonas Kubilius, Abhishek Bhandwaldar, Nick Haber, Megumi Sano,
  et~al.
\newblock Threedworld: A platform for interactive multi-modal physical
  simulation.
\newblock In {\em NeurIPS, Datasets Track}, 2021.

\bibitem{ganin2018synthesizing}
Yaroslav Ganin, Tejas Kulkarni, Igor Babuschkin, SM~Ali Eslami, and Oriol
  Vinyals.
\newblock Synthesizing programs for images using reinforced adversarial
  learning.
\newblock In {\em ICML}, 2018.

\bibitem{ganin2016domain}
Yaroslav Ganin, Evgeniya Ustinova, Hana Ajakan, Pascal Germain, Hugo
  Larochelle, Fran{\c{c}}ois Laviolette, Mario Marchand, and Victor Lempitsky.
\newblock Domain-adversarial training of neural networks.
\newblock {\em The Journal of Machine Learning Research}, 17(1):2096--2030,
  2016.

\bibitem{guo2019spottune}
Yunhui Guo, Honghui Shi, Abhishek Kumar, Kristen Grauman, Tajana Rosing, and
  Rogerio Feris.
\newblock Spottune: transfer learning through adaptive fine-tuning.
\newblock In {\em CVPR}, 2019.

\bibitem{han2021dynamic}
Yizeng Han, Gao Huang, Shiji Song, Le Yang, Honghui Wang, and Yulin Wang.
\newblock Dynamic neural networks: A survey.
\newblock {\em arXiv preprint arXiv:2102.04906}, 2021.

\bibitem{he2016deep}
Kaiming He, Xiangyu Zhang, Shaoqing Ren, and Jian Sun.
\newblock Deep residual learning for image recognition.
\newblock In {\em Proceedings of the IEEE conference on computer vision and
  pattern recognition}, pages 770--778, 2016.

\bibitem{helber2019eurosat}
Patrick Helber, Benjamin Bischke, Andreas Dengel, and Damian Borth.
\newblock Eurosat: A novel dataset and deep learning benchmark for land use and
  land cover classification.
\newblock {\em IEEE Journal of Selected Topics in Applied Earth Observations
  and Remote Sensing}, 12(7):2217--2226, 2019.

\bibitem{hinton2015distilling}
Geoffrey Hinton, Oriol Vinyals, and Jeff Dean.
\newblock Distilling the knowledge in a neural network.
\newblock {\em arXiv preprint arXiv:1503.02531}, 2015.

\bibitem{hoffman2018cycada}
Judy Hoffman, Eric Tzeng, Taesung Park, Jun-Yan Zhu, Phillip Isola, Kate
  Saenko, Alexei Efros, and Trevor Darrell.
\newblock Cycada: Cycle-consistent adversarial domain adaptation.
\newblock In {\em ICML}, 2018.

\bibitem{hull1994databaseUSPS}
Jonathan~J. Hull.
\newblock A database for handwritten text recognition research.
\newblock {\em IEEE Transactions on pattern analysis and machine intelligence},
  16(5):550--554, 1994.

\bibitem{islam2021broad}
Ashraful Islam, Chun-Fu Chen, Rameswar Panda, Leonid Karlinsky, Richard Radke,
  and Rogerio Feris.
\newblock A broad study on the transferability of visual representations with
  contrastive learning.
\newblock {\em arXiv preprint arXiv:2103.13517}, 2021.

\bibitem{johnson2017clevr}
Justin Johnson, Bharath Hariharan, Laurens Van Der~Maaten, Li Fei-Fei, C
  Lawrence~Zitnick, and Ross Girshick.
\newblock Clevr: A diagnostic dataset for compositional language and elementary
  visual reasoning.
\newblock In {\em CVPR}, 2017.

\bibitem{kar2019meta}
Amlan Kar, Aayush Prakash, Ming-Yu Liu, Eric Cameracci, Justin Yuan, Matt
  Rusiniak, David Acuna, Antonio Torralba, and Sanja Fidler.
\newblock Meta-sim: Learning to generate synthetic datasets.
\newblock In {\em ICCV}, 2019.

\bibitem{kermany2018identifyingChestXP}
Daniel~S Kermany, Michael Goldbaum, Wenjia Cai, Carolina~CS Valentim, Huiying
  Liang, Sally~L Baxter, Alex McKeown, Ge Yang, Xiaokang Wu, Fangbing Yan,
  et~al.
\newblock Identifying medical diagnoses and treatable diseases by image-based
  deep learning.
\newblock {\em Cell}, 172(5):1122--1131, 2018.

\bibitem{khirodkar2019domain}
Rawal Khirodkar, Donghyun Yoo, and Kris Kitani.
\newblock Domain randomization for scene-specific car detection and pose
  estimation.
\newblock In {\em WACV}, 2019.

\bibitem{kim2021drivegan}
Seung~Wook Kim, Jonah Philion, Antonio Torralba, and Sanja Fidler.
\newblock Drivegan: Towards a controllable high-quality neural simulation.
\newblock In {\em Proceedings of the IEEE/CVF Conference on Computer Vision and
  Pattern Recognition}, pages 5820--5829, 2021.

\bibitem{kolve2017ai2}
Eric Kolve, Roozbeh Mottaghi, Winson Han, Eli VanderBilt, Luca Weihs, Alvaro
  Herrasti, Daniel Gordon, Yuke Zhu, Abhinav Gupta, and Ali Farhadi.
\newblock Ai2-thor: An interactive 3d environment for visual ai.
\newblock {\em arXiv preprint arXiv:1712.05474}, 2017.

\bibitem{kornblith2019similarity}
Simon Kornblith, Mohammad Norouzi, Honglak Lee, and Geoffrey Hinton.
\newblock Similarity of neural network representations revisited.
\newblock In {\em International Conference on Machine Learning}, pages
  3519--3529. PMLR, 2019.

\bibitem{lake2015humanomniglot}
Brenden~M Lake, Ruslan Salakhutdinov, and Joshua~B Tenenbaum.
\newblock Human-level concept learning through probabilistic program induction.
\newblock {\em Science}, 350(6266):1332--1338, 2015.

\bibitem{langford2007epoch}
John Langford and Tong Zhang.
\newblock Epoch-greedy algorithm for multi-armed bandits with side information.
\newblock {\em Advances in Neural Information Processing Systems (NIPS 2007)},
  20:1, 2007.

\bibitem{li2017deeperPACS}
Da Li, Yongxin Yang, Yi-Zhe Song, and Timothy~M Hospedales.
\newblock Deeper, broader and artier domain generalization.
\newblock In {\em Proceedings of the IEEE international conference on computer
  vision}, pages 5542--5550, 2017.

\bibitem{little1988analysis}
James~J Little and Alessandro Verri.
\newblock Analysis of differential and matching methods for optical flow.
\newblock 1988.

\bibitem{lopez2019columnarCactusAerial}
Efren L{\'o}pez-Jim{\'e}nez, Juan~Irving Vasquez-Gomez, Miguel~Angel
  Sanchez-Acevedo, Juan~Carlos Herrera-Lozada, and Abril~Valeria Uriarte-Arcia.
\newblock Columnar cactus recognition in aerial images using a deep learning
  approach.
\newblock {\em Ecological Informatics}, 52:131--138, 2019.

\bibitem{louppe2019adversarial}
Gilles Louppe, Joeri Hermans, and Kyle Cranmer.
\newblock Adversarial variational optimization of non-differentiable
  simulators.
\newblock In {\em AISTATS}, 2019.

\bibitem{mahajan2018exploring}
Dhruv Mahajan, Ross Girshick, Vignesh Ramanathan, Kaiming He, Manohar Paluri,
  Yixuan Li, Ashwin Bharambe, and Laurens Van Der~Maaten.
\newblock Exploring the limits of weakly supervised pretraining.
\newblock In {\em Proceedings of the European conference on computer vision
  (ECCV)}, pages 181--196, 2018.

\bibitem{meng2020ar}
Yue Meng, Chung-Ching Lin, Rameswar Panda, Prasanna Sattigeri, Leonid
  Karlinsky, Aude Oliva, Kate Saenko, and Rogerio Feris.
\newblock Ar-net: Adaptive frame resolution for efficient action recognition.
\newblock In {\em ECCV}, 2020.

\bibitem{mikami2021scaling}
Hiroaki Mikami, Kenji Fukumizu, Shogo Murai, Shuji Suzuki, Yuta Kikuchi, Taiji
  Suzuki, Shin-ichi Maeda, and Kohei Hayashi.
\newblock A scaling law for synthetic-to-real transfer: How much is your
  pre-training effective?
\newblock {\em arXiv preprint arXiv:2108.11018}, 2021.

\bibitem{mohanty2016cropdisease}
Sharada~P Mohanty, David~P Hughes, and Marcel Salath{\'e}.
\newblock Using deep learning for image-based plant disease detection.
\newblock {\em Frontiers in plant science}, 7:1419, 2016.

\bibitem{netzer2011readingsvhn}
Yuval Netzer, Tao Wang, Adam Coates, Alessandro Bissacco, Bo Wu, and Andrew Ng.
\newblock Reading digits in natural images with unsupervised feature learning.
\newblock {\em NIPS Workshop on Deep Learning and Unsupervised Feature Learning
  2011}, pages 722--729, 2011.

\bibitem{nevatia1977description}
Ramakant Nevatia and Thomas~O Binford.
\newblock Description and recognition of curved objects.
\newblock {\em Artificial intelligence}, 8(1):77--98, 1977.

\bibitem{nilsback2008automatedflowers102}
Maria-Elena Nilsback and Andrew Zisserman.
\newblock Automated flower classification over a large number of classes.
\newblock In {\em 2008 Sixth Indian Conference on Computer Vision, Graphics \&
  Image Processing}, pages 722--729. IEEE, 2008.

\bibitem{oh2018self}
Junhyuk Oh, Yijie Guo, Satinder Singh, and Honglak Lee.
\newblock Self-imitation learning.
\newblock In {\em International Conference on Machine Learning}, pages
  3878--3887. PMLR, 2018.

\bibitem{DeepWeeds2019}
Alex Olsen, Dmitry~A. Konovalov, Bronson Philippa, Peter Ridd, Jake~C. Wood,
  Jamie Johns, Wesley Banks, Benjamin Girgenti, Owen Kenny, James Whinney,
  Brendan Calvert, Mostafa {Rahimi Azghadi}, and Ronald~D. White.
\newblock {DeepWeeds: A Multiclass Weed Species Image Dataset for Deep
  Learning}.
\newblock {\em Scientific Reports}, 9(2058), 2 2019.

\bibitem{peng2015learning}
Xingchao Peng, Baochen Sun, Karim Ali, and Kate Saenko.
\newblock Learning deep object detectors from 3d models.
\newblock In {\em ICCV}, 2015.

\bibitem{prakash2019structured}
Aayush Prakash, Shaad Boochoon, Mark Brophy, David Acuna, Eric Cameracci,
  Gavriel State, Omer Shapira, and Stan Birchfield.
\newblock Structured domain randomization: Bridging the reality gap by
  context-aware synthetic data.
\newblock In {\em ICRA}, 2019.

\bibitem{radford2021learning}
Alec Radford, Jong~Wook Kim, Chris Hallacy, Aditya Ramesh, Gabriel Goh,
  Sandhini Agarwal, Girish Sastry, Amanda Askell, Pamela Mishkin, Jack Clark,
  et~al.
\newblock Learning transferable visual models from natural language
  supervision.
\newblock In {\em International Conference on Machine Learning}, pages
  8748--8763. PMLR, 2021.

\bibitem{ren2018cross}
Zhongzheng Ren and Yong~Jae Lee.
\newblock Cross-domain self-supervised multi-task feature learning using
  synthetic imagery.
\newblock In {\em CVPR}, 2018.

\bibitem{rennie2017self}
Steven~J Rennie, Etienne Marcheret, Youssef Mroueh, Jerret Ross, and Vaibhava
  Goel.
\newblock Self-critical sequence training for image captioning.
\newblock In {\em Proceedings of the IEEE conference on computer vision and
  pattern recognition}, pages 7008--7024, 2017.

\bibitem{richter2021enhancing}
Stephan~R Richter, Hassan~Abu AlHaija, and Vladlen Koltun.
\newblock Enhancing photorealism enhancement.
\newblock {\em arXiv preprint arXiv:2105.04619}, 2021.

\bibitem{richter2016playing}
Stephan~R Richter, Vibhav Vineet, Stefan Roth, and Vladlen Koltun.
\newblock Playing for data: Ground truth from computer games.
\newblock In {\em European conference on computer vision}, pages 102--118.
  Springer, 2016.

\bibitem{roberto2017procedural}
Cesar Roberto~de Souza, Adrien Gaidon, Yohann Cabon, and Antonio Manuel~Lopez.
\newblock Procedural generation of videos to train deep action recognition
  networks.
\newblock In {\em CVPR}, 2017.

\bibitem{ros2016synthia}
German Ros, Laura Sellart, Joanna Materzynska, David Vazquez, and Antonio~M
  Lopez.
\newblock The synthia dataset: A large collection of synthetic images for
  semantic segmentation of urban scenes.
\newblock In {\em CVPR}, 2016.

\bibitem{rozantsev2018beyond}
Artem Rozantsev, Mathieu Salzmann, and Pascal Fua.
\newblock Beyond sharing weights for deep domain adaptation.
\newblock {\em IEEE transactions on pattern analysis and machine intelligence},
  41(4), 2018.

\bibitem{ruiz2018learning}
Nataniel Ruiz, Samuel Schulter, and Manmohan Chandraker.
\newblock Learning to simulate.
\newblock In {\em ICLR}, 2019.

\bibitem{savva2019habitat}
Manolis Savva, Abhishek Kadian, Oleksandr Maksymets, Yili Zhao, Erik Wijmans,
  Bhavana Jain, Julian Straub, Jia Liu, Vladlen Koltun, Jitendra Malik, et~al.
\newblock Habitat: A platform for embodied ai research.
\newblock In {\em ICCV}, 2019.

\bibitem{shrivastava2017learning}
Ashish Shrivastava, Tomas Pfister, Oncel Tuzel, Joshua Susskind, Wenda Wang,
  and Russell Webb.
\newblock Learning from simulated and unsupervised images through adversarial
  training.
\newblock In {\em CVPR}, 2017.

\bibitem{su2015render}
Hao Su, Charles~R Qi, Yangyan Li, and Leonidas~J Guibas.
\newblock Render for cnn: Viewpoint estimation in images using cnns trained
  with rendered 3d model views.
\newblock In {\em Proceedings of the IEEE International Conference on Computer
  Vision}, pages 2686--2694, 2015.

\bibitem{sun2017revisiting}
Chen Sun, Abhinav Shrivastava, Saurabh Singh, and Abhinav Gupta.
\newblock Revisiting unreasonable effectiveness of data in deep learning era.
\newblock In {\em Proceedings of the IEEE international conference on computer
  vision}, pages 843--852, 2017.

\bibitem{sun2019adashare}
Ximeng Sun, Rameswar Panda, Rogerio Feris, and Kate Saenko.
\newblock Adashare: Learning what to share for efficient deep multi-task
  learning.
\newblock In {\em NeurIPS}, 2020.

\bibitem{tian2020kaokore}
Yingtao Tian, Chikahiko Suzuki, Tarin Clanuwat, Mikel Bober-Irizar, Alex Lamb,
  and Asanobu Kitamoto.
\newblock Kaokore: A pre-modern japanese art facial expression dataset.
\newblock {\em arXiv preprint arXiv:2002.08595}, 2020.

\bibitem{tobin2017domain}
Josh Tobin, Rachel Fong, Alex Ray, Jonas Schneider, Wojciech Zaremba, and
  Pieter Abbeel.
\newblock Domain randomization for transferring deep neural networks from
  simulation to the real world.
\newblock In {\em IROS}, 2017.

\bibitem{tzeng2017adversarial}
Eric Tzeng, Judy Hoffman, Kate Saenko, and Trevor Darrell.
\newblock Adversarial discriminative domain adaptation.
\newblock In {\em CVPR}, 2017.

\bibitem{varol2021synthetic}
G{\"u}l Varol, Ivan Laptev, Cordelia Schmid, and Andrew Zisserman.
\newblock Synthetic humans for action recognition from unseen viewpoints.
\newblock {\em International Journal of Computer Vision}, 129(7):2264--2287,
  2021.

\bibitem{veit2018convolutional}
Andreas Veit and Serge Belongie.
\newblock Convolutional networks with adaptive inference graphs.
\newblock In {\em ECCV}, 2018.

\bibitem{WahCUB_200_2011}
C. Wah, S. Branson, P. Welinder, P. Perona, and S. Belongie.
\newblock {The Caltech-UCSD Birds-200-2011 Dataset}.
\newblock Technical report, 2011.

\bibitem{wang2019learningsketch}
Haohan Wang, Songwei Ge, Eric~P. Xing, and Zachary~C. Lipton.
\newblock Learning robust global representations by penalizing local predictive
  power.
\newblock {\em arXiv preprint arXiv:1905.13549}, 2019.

\bibitem{wang2017chestx}
Xiaosong Wang, Yifan Peng, Le Lu, Zhiyong Lu, Mohammadhadi Bagheri, and
  Ronald~M Summers.
\newblock Chestx-ray8: Hospital-scale chest x-ray database and benchmarks on
  weakly-supervised classification and localization of common thorax diseases.
\newblock In {\em Proceedings of the IEEE Conference on Computer Vision and
  Pattern Recognition}, pages 2097--2106, 2017.

\bibitem{wang2018skipnet}
Xin Wang, Fisher Yu, Zi-Yi Dou, Trevor Darrell, and Joseph~E Gonzalez.
\newblock Skipnet: Learning dynamic routing in convolutional networks.
\newblock In {\em ECCV}, 2018.

\bibitem{wang2020differential}
Zhonghao Wang, Mo Yu, Yunchao Wei, Rogerio Feris, Jinjun Xiong, Wen-mei Hwu,
  Thomas~S Huang, and Honghui Shi.
\newblock Differential treatment for stuff and things: A simple unsupervised
  domain adaptation method for semantic segmentation.
\newblock In {\em CVPR}, 2020.

\bibitem{rw2019timm}
Ross Wightman.
\newblock Pytorch image models.
\newblock \url{https://github.com/rwightman/pytorch-image-models}, 2019.

\bibitem{williams1992simple}
Ronald~J Williams.
\newblock Simple statistical gradient-following algorithms for connectionist
  reinforcement learning.
\newblock {\em Machine learning}, 8(3):229--256, 1992.

\bibitem{wu2018blockdrop}
Zuxuan Wu, Tushar Nagarajan, Abhishek Kumar, Steven Rennie, Larry~S Davis,
  Kristen Grauman, and Rogerio Feris.
\newblock Blockdrop: Dynamic inference paths in residual networks.
\newblock In {\em CVPR}, 2018.

\bibitem{wu2019adaframe}
Zuxuan Wu, Caiming Xiong, Chih-Yao Ma, Richard Socher, and Larry~S Davis.
\newblock Adaframe: Adaptive frame selection for fast video recognition.
\newblock In {\em CVPR}, 2019.

\bibitem{xia2018gibson}
Fei Xia, Amir~R Zamir, Zhiyang He, Alexander Sax, Jitendra Malik, and Silvio
  Savarese.
\newblock Gibson env: Real-world perception for embodied agents.
\newblock In {\em CVPR}, 2018.

\bibitem{xia2017aid}
Gui-Song Xia, Jingwen Hu, Fan Hu, Baoguang Shi, Xiang Bai, Yanfei Zhong,
  Liangpei Zhang, and Xiaoqiang Lu.
\newblock Aid: A benchmark data set for performance evaluation of aerial scene
  classification.
\newblock {\em IEEE Transactions on Geoscience and Remote Sensing},
  55(7):3965--3981, 2017.

\bibitem{yang2020learning}
Dawei Yang and Jia Deng.
\newblock Learning to generate 3d training data through hybrid gradient.
\newblock In {\em Proceedings of the IEEE/CVF Conference on Computer Vision and
  Pattern Recognition}, pages 779--789, 2020.

\bibitem{yue2019domain}
Xiangyu Yue, Yang Zhang, Sicheng Zhao, Alberto Sangiovanni-Vincentelli, Kurt
  Keutzer, and Boqing Gong.
\newblock Domain randomization and pyramid consistency: Simulation-to-real
  generalization without accessing target domain data.
\newblock In {\em ICCV}, 2019.

\bibitem{yun2019cutmix}
Sangdoo Yun, Dongyoon Han, Seong~Joon Oh, Sanghyuk Chun, Junsuk Choe, and
  Youngjoon Yoo.
\newblock Cutmix: Regularization strategy to train strong classifiers with
  localizable features.
\newblock In {\em Proceedings of the IEEE/CVF International Conference on
  Computer Vision}, pages 6023--6032, 2019.

\bibitem{zhang2017mixup}
Hongyi Zhang, Moustapha Cisse, Yann~N Dauphin, and David Lopez-Paz.
\newblock mixup: Beyond empirical risk minimization.
\newblock {\em arXiv preprint arXiv:1710.09412}, 2017.

\bibitem{zhang2019curriculum}
Yang Zhang, Philip David, Hassan Foroosh, and Boqing Gong.
\newblock A curriculum domain adaptation approach to the semantic segmentation
  of urban scenes.
\newblock {\em IEEE transactions on pattern analysis and machine intelligence},
  42(8):1823--1841, 2019.

\bibitem{zhang2019poissonFMD}
Yide Zhang, Yinhao Zhu, Evan Nichols, Qingfei Wang, Siyuan Zhang, Cody Smith,
  and Scott Howard.
\newblock A poisson-gaussian denoising dataset with real fluorescence
  microscopy images.
\newblock In {\em Proceedings of the IEEE/CVF Conference on Computer Vision and
  Pattern Recognition}, pages 11710--11718, 2019.

\end{thebibliography}
